\pdfoutput=1

\documentclass[11pt]{article}

\usepackage[final]{acl}

\usepackage{times}
\usepackage{latexsym}

\usepackage[T1]{fontenc}

\usepackage[utf8]{inputenc}

\usepackage{microtype}

\usepackage{inconsolata}

\usepackage{graphicx}
\usepackage{algorithm}
\usepackage{algorithmic}
\usepackage{booktabs}
\usepackage{makecell}
\usepackage{bm}
\usepackage{pifont}
\usepackage{hyperref}
\usepackage{amsmath}
\usepackage{amsthm}
\usepackage{amsfonts}
\usepackage{multirow} 
\usepackage{color}

\usepackage{newfloat}
\usepackage{listings}

\definecolor{lightgreen}{rgb}{0.55, 0.71, 0.0}
\definecolor{bisque}{rgb}{0.87, 0.72, 0.53}
\definecolor{lightyellow}{rgb}{0.99, 0.76, 0.0}
\definecolor{lightblue}{rgb}{0.36, 0.54, 0.66}
\definecolor{darkgray}{rgb}{0.66, 0.66, 0.66}
\definecolor{salmon}{rgb}{0.98, 0.50, 0.45}
\definecolor{deeppurple}{rgb}{0.4, 0.0, 0.4}

\usepackage[switch]{lineno}

\title{Eliminating Biased Length Reliance of Direct Preference Optimization via Down-Sampled KL Divergence}

\author{\makecell{Junru Lu$^{1*,3}$, Jiazheng Li$^2$\thanks{Equal Contribution.}, Siyu An$^3$, Meng Zhao$^3$, Yulan He$^{1,2,4}$, Di Yin$^3$, Xing Sun$^3$} \\
  $^1$University of Warwick\quad\quad $^2$King's College London \\$^3$Tencent YouTu Lab\quad\quad $^4$The Alan Turing Institute\\
    \texttt{junru.lu@warwick.ac.uk}, \texttt{\{jiazheng.li, yulan.he\}@kcl.ac.uk}\\
  \texttt{\{siyuan, alexmzhao, endymecyyin, winfredsun\}@tencent.com}}

\begin{document}
\maketitle
\begin{abstract}
Direct Preference Optimization (DPO) has emerged as a prominent algorithm for the direct and robust alignment of Large Language Models (LLMs) with human preferences, offering a more straightforward alternative to the complex Reinforcement Learning from Human Feedback (RLHF). Despite its promising efficacy, DPO faces a notable drawback: ``\emph{verbosity}'', a common over-optimization phenomenon also observed in RLHF. While previous studies mainly attributed verbosity to biased labels within the data, we propose that the issue also stems from an inherent algorithmic length reliance in DPO. Specifically, we suggest that the discrepancy between sequence-level Kullback–Leibler (KL) divergences between chosen and rejected sequences, used in DPO, results in overestimated or underestimated rewards due to varying token lengths. Empirically, we utilize datasets with different label lengths to demonstrate the presence of biased rewards. We then introduce an effective downsampling approach, named SamPO, to eliminate potential length reliance. Our experimental evaluations, conducted across three LLMs of varying scales and a diverse array of conditional and open-ended benchmarks, highlight the efficacy of SamPO in mitigating verbosity, achieving improvements of 5\% to 12\% over DPO through debaised rewards\footnote{Our code can be accessed at: \url{https://github.com/LuJunru/SamPO/}.}.
\end{abstract}

\section{Introduction}
Reinforcement Learning from Human Feedback (RLHF) is a crucial strategy for effectively align Large Language Models (LLMs) with human minds\,\cite{zhao2023survey,yang2023harnessing,pan2023automatically}, showcasing significant improvements of LLM's instruct-following capability compared with the other two popular approaches: pre-training and supervised fine-tuning (SFT). In fact, a series of leading LLMs have adopted RLHF as the final stage of their entire training pipelines\,\cite{ouyang2022training,openai2023gpt4,bi2024deepseek}.

Nevertheless, traditional RLHF involves several intricate multi-stage steps, typically starting with fine-tuning a reward model that captures complex human intuition\,\cite{bai2022training}, followed by optimizing LLMs to maximize preference scores. Therefore, the quality of the reward model is crucial. However, modeling elusive human intuition is inherently difficult\,\cite{wang2024secrets}. On the contrary, Direct Preference Optimization (DPO)\,\cite{rafailov2023direct} proposed to re-parameterize the reward model, integrating preference feedback from online rewards into offline labels. In specific, DPO employs the Bradley-Terry model\,\cite{bradley1952rank} to maximize implicit rewards via pairwise offline preference labels. The implicit reward is mathematically equivalent to the discrepancy in sequence-level Kullback–Leibler (KL) divergences\,\cite{kullback1951information} between chosen and rejected labels. The KL divergence for each label is calculated based on probability outputs from the fine-tuning policy model and a frozen reference model. DPO eliminates the need for complex prefix fine-tuning of an external reward model, while maintains performance comparable to RLHF\,\cite{dubois2024alpacafarm,hou2024chatglm}.

\begin{figure*}[!t]
  \centering
  \includegraphics[width=1.0\linewidth]{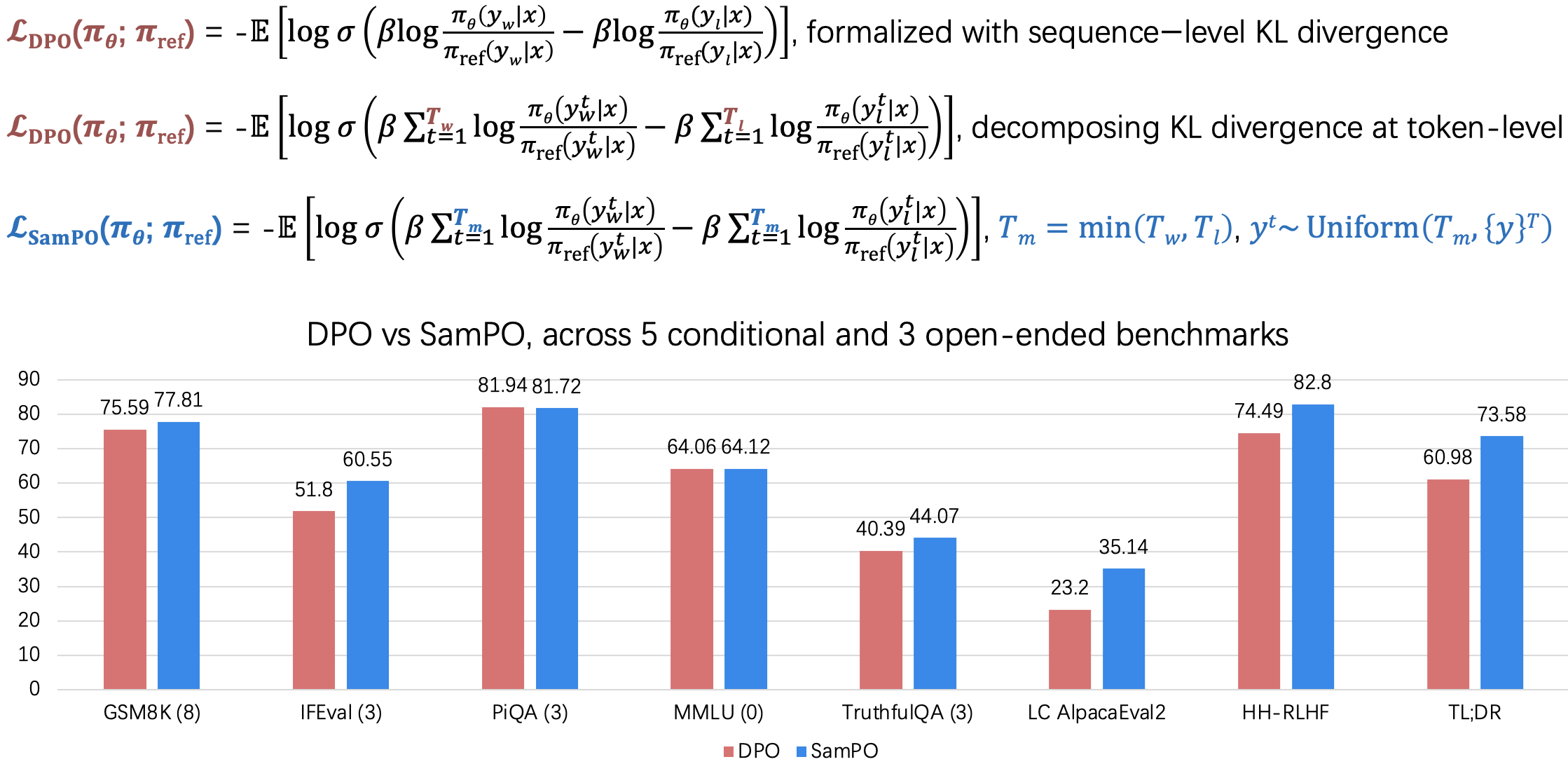}
  \caption{Down-Sampling strategy helps mitigate the potential length reliance, and thus improves DPO.}
  \label{fig:intro}
\end{figure*}

Despite its effectiveness, DPO faces several notable challenges, including issues of overfitting\,\cite{azar2023general,jung2024binary}, high computational costs\,\cite{ethayarajh2024kto,hong2024reference}, and verbosity\,\cite{hou2024chatglm,park2024disentangling}. This paper specifically focuses on addressing the ``\emph{verbosity}'' issue. 

Traditional multi-stage RLHF methods argue that due to a statistical bias in length distribution, that is, where  preferred labels tend to be longer than  rejected preference labels\,\cite{singhal2023long,park2024disentangling}, the reward model trained on such preference data  inherently exhibit a length bias\,\cite{shen2023loose}. Therefore, subsequent fine-tuned policy model exploit this bias as a shortcut to achieve higher reward scores by generating longer responses\,\cite{gao2023scaling}, without necessarily improving  quality\,\cite{kabir2023answers,dubois2024alpacafarm}. Various regularization approaches have been proposed to mitigate this inherent bias within reward models\,\cite{ramamurthy2022reinforcement,coste2023reward,moskovitz2023confronting,chen2024odin}. On the other hand, although DPO does not explicitly use a reward model, the length distribution bias inherent in the offline preference labels still contributes to the verbosity issue\,\cite{hou2024chatglm,rafailov2024scaling}. Analysis suggests that policy models trained using DPO tend to generate responses that are almost twice the length of the labeled data\,\cite{park2024disentangling}.

In this paper, we propose that, in addition to the length bias in the data, DPO exhibits a hidden algorithmic dependence on response length. As illustrated in the upper portion of Figure\,\ref{fig:intro}, the loss function in DPO is based on the discrepancy between sequence-level KL divergence, which can also be computed and aggregated at the token-level. It is evident that discrepancies between chosen label $\bm{y}_w$ and rejected label $\bm{y}_l$ lead to an inadvertent reliance on auxiliary length features: training samples with longer chosen labels than rejected ones lead to overestimated rewards during training, while those with shorter chosen labels result in underestimated rewards. Therefore, overestimated rewards contribute more significantly to gradient optimization, ultimately exacerbating verbosity. We believe this algorithmic dependence on response length is a unique drawback of DPO, since the explicit rewards in RLHF typically manifest as scalar values\,\cite{ouyang2022training}. 

We propose that addressing this reliance on response length can be effectively achieved through a straightforward down-sampling method. Illustrated in the middle of Figure\,\ref{fig:intro}, this approach involves down-sampling equal token-level probability features for computing regularized KL divergences. Our contributions in this paper are threefold:
\begin{itemize}
    \item We analyze the algorithmic dependence on response length in DPO, revaling how it results in overestimated or underestimated rewards. Through decomposition experiments using datasets with varying label length, we empirically demonstrate the biased rewards.
    \item We propose a lightweight approach, called SamPO, to mitigate the biased length reliance in DPO. By simply down-sampling equal probability features at the token-level, we can apply DPO with regularized KL divergences.
    \item We validate our method using three different LLMs of varying scales. Compared to DPO, SamPO significantly reduces verbosity. Leveraging debaised rewards, we achieve significant  improvements across five conditioned and three open-ended benchmarks, as depicted in the lower section of Figure\,\ref{fig:intro}.
\end{itemize}

\section{Related Work}

\noindent \textbf{Optimization from Human Preference} aims to align neural models with human minds. As a seminal work, 
\,\cite{stiennon2020learning} collected human preferences on 123k pairs of summary outputs, then trained a reward model that guides the GPT-3 model\,\cite{brown2020language} to produce more coherent and human-preferred summaries. \,\cite{ouyang2022training} then further scaled similar pipeline with 1M diverse text instructions, and reported that outputs from the 1.3B parameter InstructGPT model were preferred to outputs from the 175B GPT-3 model, according to downstream human evaluation. RLHF has become an essential part of aligning LLMs\,\cite{touvron2023llama,bi2024deepseek,bai2023qwen,young2024yi}. However, as it follows a multi-stage training strategy, and heavily relays on the quality of reward model, RLHF's training cost and stability are widely criticized\,\cite{zheng2023secrets,mckinney2023fragility}. Therefore, DPO came into being, providing a stable alternative that does not rely on an explicit reward model\,\cite{rafailov2023direct}. It has been proved that DPO can achieve the same alignment effect as RLHF\,\cite{ivison2023camels,hou2024chatglm}. 

\noindent \textbf{Over-optimization in RL} is a well-known obstacle\,\cite{skalse2022defining,pan2023rewards,casper2023open,zheng2023secrets}, which refers to the phenomenon that feedback scores from the reward model are getting higher, but the updated policy model produces lower quality responses. And one particularly noticeable low-quality feature is verbosity. It is general to blame for exploitation of reward model\,\cite{casper2023open,gao2023scaling}, and thus various regularization approaches have been proposed, including uncertainty-based regularization\,\cite{coste2023reward,zhai2023uncertainty}, composite reward models\,\cite{moskovitz2023confronting}, and length decorrelation\,\cite{chen2024odin}. However, since the reward model is eliminated in DPO, none of the above approaches can be directly applied. Herein, specific methods are introduced, \,\cite{park2024disentangling} introduced a pairwise length regularization term to dampen the verbosity trends, and SimPO\,\cite{meng2024simpo} used average probability to eliminate length reliance.

In this paper, we present that the verbosity issue in DPO is further related to algorithmic biased length reliance, which is never analyzed in previous literature. And this drawback can be effectively handled via down-sampling over KL divergence.

\section{SamPO: Down-Sampled DPO}
In this section, we first give a brief introduction of DPO's optimization target (\S\ref{sub:dpo}), then dive into further analysis of its potential length reliance (\S\ref{sub:length}). Subsequently, we present SamPO, which intuitively regularizes the biased length-specific reward (\S\ref{sub:sampo}).

\begin{figure*}[!t]
  \centering
  \includegraphics[width=1.0\linewidth]{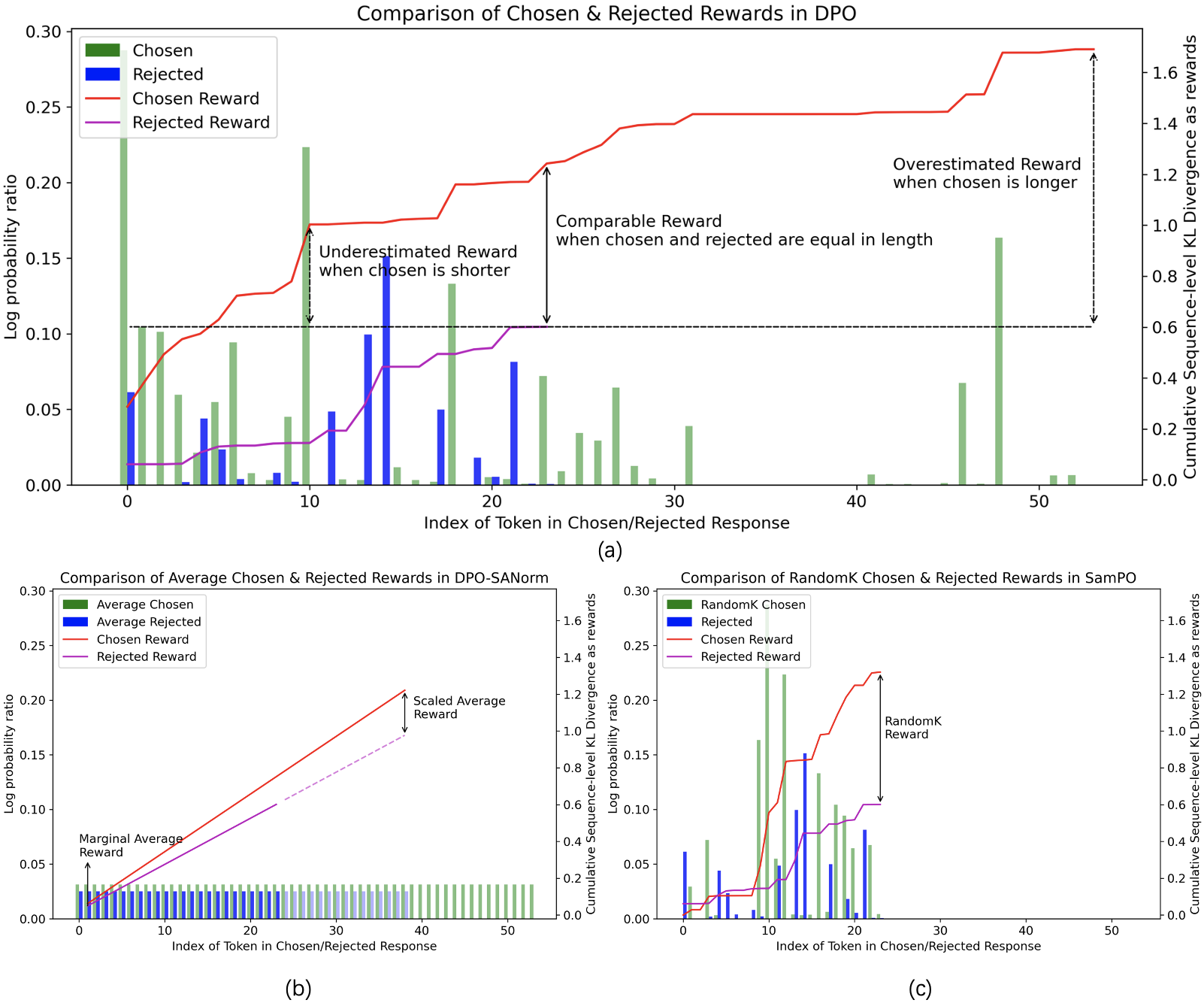}
  \caption{The inequality of pairwise responses, illustrated by typical examples, forces DPO to overestimate or underestimate actual rewards. In the upper sub-figure (a), we present DPO's chosen reward $\sum\log\frac{\pi_{\theta}(y_w|x)}{\pi_{ref}(y_w|x)}$ and rejected reward $\sum\log\frac{\pi_{\theta}(y_l|x)}{\pi_{ref}(y_l|x)}$ with {\color{red}{red}} and {\color{deeppurple}{purple}} curves, respectively. Each response's reward is sequence-level KL divergence, summarizing from token-level log probability ratios (the {\color{lightgreen}{green}} and {\color{blue}{blue}} bars). Therefore, the difference gap between two lines draws the implicit reward target in DPO, as shown in Eq.\,\ref{eqa:dpo_2_t}. Similarly, we show averagely normalized DPO in left bottom sub-figure (b), and show our SamPO in right bottom sub-figure (c).}
  \label{fig:rewards}
\end{figure*}

\subsection{Preliminary Background of DPO}
\label{sub:dpo}
DPO implements direct RLHF based on offline preference data and offloaded reward model. Specifically, DPO first re-parameterizes the reward model in multi-stage RLHF as follows:
\begin{equation}
  \label{eqa:dpo_r}
  r_\phi(x,y)=\beta \log\frac{\pi_{\theta}(y|x)}{\pi_{ref}(y|x)}+\beta \log Z(x)
\end{equation}
where $\bm{r}_\phi$, $\bm{\pi}_{\theta}$ and $\bm{\pi}_{ref}$ stand for reward model, policy model and reference model, respectively. $\bm{\pi}_{\theta}$ and $\bm{\pi}_{ref}$ are usually initialized from same SFT model. $\bm{\pi}_{\theta}$ is about to be further optimized in DPO, while $\bm{\pi}_{ref}$ is usually frozen. $\bm{Z(x)}$ is the partition function, and $\bm{\beta}$ is a hyperparameter that adjust the intensity of rewards. DPO then incorporates Bradley-Terry model:
\begin{equation}\small
  \label{eqa:bt_model}
  P_\theta(y_w \succ y_l | x) = \frac{\exp(r_\phi(x, y_w))}{\exp(r_\phi(x, y_w)) + \exp(r_\phi(x, y_l))}
\end{equation}
where a preference triplet $(\bm{x}, \bm{y}_w, \bm{y}_l)$ consists of a prompt instruction $\bm{x}$, a chosen response $\bm{y}_w$, and a rejected response $\bm{y}_l$. The Bradley-Terry model hypothesizes that preference probability $\bm{P}_\theta$ can be estimated via pairwise comparison.
\begin{equation}
  \label{eqa:dpo_1}
  \mathcal{L}_{dpo}(\pi_{\theta};\pi_{ref}) = -\mathbb{E}_{(x, y_w, y_l) \sim D}[\log\sigma(\Delta)]
\end{equation}
\begin{equation}
  \label{eqa:dpo_2}
  \Delta = \beta \log\frac{\pi_{\theta}(y_w|x)}{\pi_{ref}(y_w|x)}-\beta \log\frac{\pi_{\theta}(y_l|x)}{\pi_{ref}(y_l|x)}
\end{equation}
where Eq.\,\ref{eqa:dpo_1} stands for the loss of DPO by inserting Eq.\,\ref{eqa:dpo_r} into Eq.\,\ref{eqa:bt_model}. $\bm{\sigma}$ stands for sigmoid function, and $\bm{D}$ denotes the entire pairwise preference dataset. The implicit reward $\bm{\Delta}$ in Eq.\,\ref{eqa:dpo_2} is formalized as discrepancy between chosen KL divergence $\log\frac{\pi_{\theta}(y_w|x)}{\pi_{ref}(y_w|x)}$ and rejected KL divergence $\log\frac{\pi_{\theta}(y_l|x)}{\pi_{ref}(y_l|x)}$. Each KL divergence is calculated upon the tokens in the response $\bm{y}$. Considering Eq.\,\ref{eqa:dpo_1}, the gradients of DPO can be written as:
\begin{equation}
  \label{eqa:dpo_g_1}\small
  \nabla_{\theta}\mathcal{L}_{dpo}(\pi_{\theta};\pi_{ref}) = -\mathbb{E}_{(x, y_w, y_l) \sim D}[\beta\sigma(-\Delta)\mathcal{M}]
\end{equation}
\begin{equation}
  \label{eqa:dpo_g_2}
  \mathcal{M} = \nabla_{\theta}\log\pi(y_w|x)-\nabla_{\theta}\log\pi(y_l|x)
\end{equation}
where $\bm{\mathcal{M}}$ is a discrepancy term that lead the policy model $\bm{\pi}_\theta$ to increase likelihood of chosen response $\bm{y}_w$ and decrease likelihood of rejected response $\bm{y}_l$ simultaneously. The $\bm{\Delta}$ is regarded as the weight that scales the intensity of $\bm{\mathcal{M}}$.

\subsection{Biased Length Reliance in DPO}
\label{sub:length}
The loss of DPO and its gradient are computed at sequence-level. That is, when calculating the KL divergence $\log\frac{\pi_{\theta}(y|x)}{\pi_{ref}(y|x)}$, DPO treats probabilities of tokens as discrete samples. Therefore, we can denote Eq.\,\ref{eqa:dpo_2} at token-level (Proof in App.\,\ref{sec:app_proof}):
\begin{equation}
  \label{eqa:dpo_2_t}\small
  \Delta = \beta \sum\limits_{t=1}^{T_w} \log\frac{\pi_{\theta}(y_w^t|x)}{\pi_{ref}(y_w^t|x)}-\beta \sum\limits_{t=1}^{T_l}\log\frac{\pi_{\theta}(y_l^t|x)}{\pi_{ref}(y_l^t|x)}
\end{equation}
where $\bm{T}_w$ and $\bm{T}_l$ demote the number of 1st to $\bm{t}$-th tokens in chosen response $\bm{y}_w$ and rejected response $\bm{y}_l$, respectively. Similarly, we re-write Eq.\,\ref{eqa:dpo_g_2}:
\begin{equation}
  \label{eqa:dpo_g_2_t}\small
  \mathcal{M} = \nabla_{\theta}\sum\limits_{t=1}^{T_w}\log\pi(y_w^t|x)-\nabla_{\theta}\sum\limits_{t=1}^{T_l}\log\pi(y_l^t|x)
\end{equation}

Herein, we can intuitively perceive the impact of the inequality in length between chosen response $\bm{y}_w$ and rejected response $\bm{y}_l$ on loss and gradient. As illustrated in sub-Figure\,\ref{fig:rewards}(a), a ``\emph{comparable reward}'' is obtained if $\bm{y}_w$ and $\bm{y}_l$ share same length, with which DPO fairly learns the quality difference. However, if $\bm{y}_w$ is much longer than $\bm{y}_l$, it is likely that the significant lead in the total amount of tokens in $\bm{y}_w$ brings about ``\emph{overestimated reward}'' in Eq.\,\ref{eqa:dpo_2_t}, and contributes more to the gradient updates in Eq.\,\ref{eqa:dpo_g_1} and\,\ref{eqa:dpo_g_2_t}. Reversely, DPO could ``\emph{underestimate reward}'' and adopt less gradients if $\bm{y}_w$ is shorter than $\bm{y}_l$, even $\bm{y}_w$ has better quality. This biased length reliance makes DPO prefer long, plausible chosen responses but disregards short, sound chosen responses in training, leading to verbosity.

\subsection{Debiased KL Divergence}
\label{sub:sampo}
We consider two common strategies to eliminate length reliance: averaging and sampling.

\textbf{Averaging} the sequence-level KL divergence to instead use marginal averaged reward is the most straightforward idea for length regularization, which modify Eq.\,\ref{eqa:dpo_2_t} as follows:
\begin{equation}
  \label{eqa:dpo_2_t_a}\small
  \Delta = \beta \frac{\sum\limits_{t=1}^{T_w} \log\frac{\pi_{\theta}(y_w^t|x)}{\pi_{ref}(y_w^t|x)}}{|T_w|} - \beta \frac{\sum\limits_{t=1}^{|T_l|}\log\frac{\pi_{\theta}(y_l^t|x)}{\pi_{ref}(y_l^t|x)}}{|T_l|}
\end{equation}
Such averaged normalization can simply eliminate length reliance. However, as shown in the left corner of Figure\,\ref{fig:rewards}(b), there lies scale difference between marginal averaged reward with original sequence-level reward. Thus, we scale the marginal reward with a dynamic scaling factor $\frac{(T_w + T_l)}{2}$, which is the average length of chosen response $\bm{y}_w$ and rejected response $\bm{y}_l$.

\textbf{Sampling} same amount of tokens from the chosen and rejected responses, and calculating down-sampled sequence-level KL divergence for the implicit reward. Eq.\,\ref{eqa:dpo_2_t} as follows:
\begin{equation}
  \label{eqa:dpo_2_t_s}\small
  \begin{aligned}
  \Delta &= \beta \sum\limits_{t=1}^{T_m} \log\frac{\pi_{\theta}(y_w^t|x)}{\pi_{ref}(y_w^t|x)}-\beta \sum\limits_{t=1}^{T_m}\log\frac{\pi_{\theta}(y_l^t|x)}{\pi_{ref}(y_l^t|x)}\\
  &T_m=\mbox{min}(T_w, T_l)\mbox{, }y^t\sim~\mbox{Uniform}(T_m, \{y\}^T)
  \end{aligned}
\end{equation}
where $\bm{T}_m$ equals to the minimum token length of $(\bm{T}_w, \bm{T}_l)$, and $\bm{y}^t$ is down-sampled from all tokens $\{\bm{y}^T\}$ uniformly. Eq.\,\ref{eqa:dpo_2_t_s} is consistent with the corresponding reward term in Figure\,\ref{fig:intro}, as shown in the middle. In addition, we provide ablation studies of sampling randomness in Appendix\,\ref{sec:app_seed}.

Figure\,\ref{fig:rewards}(b) and (c) demonstrate that either averaging or sampling can lead to comparable length-debiased rewards. However, the simple averaging vanishes the variance feature among tokens. We thus confirm to use down-sampling strategy in proposed SamPO. We validate this choice in section\,\ref{sec:exp_result}.

\section{Experimental Setup}
\label{sec:exp_setup}
In this section, we begin with our datasets (\S\,\ref{sec:exp_dataset}, \S\,\ref{sec:exp_benchmark}), then follow with baselines (\S\,\ref{sec:exp_llm}, \S\,\ref{sec:exp_baseline}), and an overview of our experimental design (\S\,\ref{sec:exp_design}).

\subsection{Training Datasets}
\label{sec:exp_dataset}
We include three independent preference datasets for training. Two datasets in consistent with the original DPO\,\cite{rafailov2023direct}: The 161k HH-RLHF data\,\cite{ganguli2022red}, and the 92.8k TL;DR data\,\cite{volske-etal-2017-tl}. And the 61k binarized UltraFeedback data\,\cite{cui2023ultrafeedback} that has been adopted in many following works\,\cite{ivison2023camels,meng2024simpo} of DPO. All datasets own evaluation sets for cross-validation in training.

\subsection{Evaluation Benchmarks}
\label{sec:exp_benchmark}
Following DPO, with models training upon HH-RLHF or TL;DR, we randomly select 256 samples from its evaluation set for final testing. We report the win rate between the response of fine-tuned policy model $\hat{\bm{y}_{\theta}}$ = $\bm{\pi}_{\theta}(\bm{x}_{test})$ and the response of basis SFT model $\hat{\bm{y}_{ref}}$ = $\bm{\pi}_{ref}(\bm{x}_{test})$, judged by GPT-4\,\cite{openai2023gpt4}. As for models training with UltraFeedback, we use five conditional and one open-ended generation benchmarks. The conditional benchmarks and their in-context examples are: GSM8K in 8-shot\,\cite{cobbe2021training}, IFEval in 3-shot\,\cite{zhou2023instructionfollowing}, PiQA in 3-shot\,\cite{bisk2020piqa}, MMLU in 0-shot\,\cite{hendrycks2021measuring}, and TruthfulQA in 3-shot\,\cite{lin-etal-2022-truthfulqa}. The open-ended data is AlpacaEval2\,\cite{alpaca_eval}. We report match accuracy of conditional benchmarks, and report length-debiased GPT-4 win rate of AlpacaEval2\,\cite{dubois2024length}. Check more evaluation details in Appendix\,\ref{sec:app_eval}.

\subsection{Foundation Models}
\label{sec:exp_llm}
We include LLMs varying three different scales: Pythia-2.8B\,\cite{biderman2023pythia}, Llama3-8B-Instruct\,\cite{llama3modelcard}, and Tulu2-13B-SFT \,\cite{ivison2023camels}. We report details of these LLMs, hyperparameters, and cost in Appendix\,\ref{sec:app_cost}.

\subsection{Baselines}
\label{sec:exp_baseline}
Many variants of DPO have been proposed, and can be categorized into three types: (1) \textbf{Reduce cost}. Despite the robustness, preparing high-quality pair-wise preference labels and running with two giant models makes DPO still expensive. As such, KTO\,\cite{ethayarajh2024kto} proposed to use non-pairwise preference data. ORPO\,\cite{hong2024reference}, CPO\,\cite{xu2024contrastive}, and SimPO\,\cite{meng2024simpo} suggested new reference-free losses to optimize with single policy model; (2) \textbf{Alleviate overfitting}. IPO\,\cite{azar2023general} analyzed the risk of overfitting, and presented a square loss to reform the monotonous DPO loss. TDPO\,\cite{zeng2024token} incorporated forward KL divergence constraints for each token, improving alignment and diversity. BCO\,\cite{jung2024binary} and NCA\,\cite{chen2024noise} provided plans to relieve the noise from the pairwise preference responses; (3) \textbf{Overcome verbosity}. One notable degradation side effect of DPO is verbosity, which refers to generate longer and longer responses during training, while the actual answer quality are not improved at the same time. \,\cite{park2024disentangling} introduced a pairwise length regularization term to dampen the verbosity trends, and SimPO\,\cite{meng2024simpo} used average probability to eliminate length reliance.

We select methods that are targeted at noise removal or length normalization and have relatively positive testing results as our final baselines: Hybrid DPO+SFT, TDPO\,\cite{zeng2024token}, Length-normed DPO\,\cite{park2024disentangling}, BCO\,\cite{jung2024binary}, SimPO\,\cite{meng2024simpo}. Particularly, Hybrid DPO+SFT refers to multi-task learning of DPO on the pairwise responses and SFT on the chosen response at the same time, which is a common practice\,\cite{hua2024intuitive,lu2024online}.

\subsection{Experimental Designs}
\label{sec:exp_design}
In general, we design three groups of experiments: 
\begin{itemize}
    \item[(1)] \textbf{Presence of biased length reliance}. We extract two 27k subsets from the UltraFeedback only by response length. One is named UltraFeedback-long, in which the chosen response of each data must be longer than the rejected response. The other one is named UltraFeedback-short, and as the name suggests, it contains shorter chosen response. We use these subsets for biased reward exhibition.
    \item[(2)] \textbf{Preliminary Study of DPO and variants}. Given that there are many variants of DPO, and they often use their own hyperparameters, we first conduct a preliminary study to align their performance under the same conditions. This study helps us select several robust baselines. The results are reported in Appendix\,\ref{sec:app_trail}.
    \item[(3)] \textbf{Experiments with various LLMs}. Similar to DPO, we use Pythia-2.8B to train and test SamPO on HH-RLHF or TL;DR; on the other hand, following relevant studies\,\cite{ivison2023camels,hong2024reference}, we use Tulu2-13B-SFT and Llama3-8B-Instruct to train on Ultrafeedback and verify SamPO on public benchmarks. Also, literature reports that iteratively update the frozen reference model $\bm{\pi}_{ref}$ can obtain further gains\,\cite{gorbatovski2024learn,zhang2024map}. Thus, we combine it with SamPO to present Iterative SamPO.
\end{itemize}

\section{Experimental Results}
\label{sec:exp_result}
In this section, following above designs, we first report the group experiments of length reliance (\S\,\ref{sec:exp_group}), then present comparison studies against strong baselines (\S\,\ref{sec:exp_comparison}). We discuss quantitative results in the main body. We leave more ablation studies and case analysis in Appendix\,\ref{sec:app_seed}, \,\ref{sec:app_beta}, and \,\ref{sec:app_case}.

\begin{figure}[!t]
  \centering
  \includegraphics[width=1.0\linewidth]{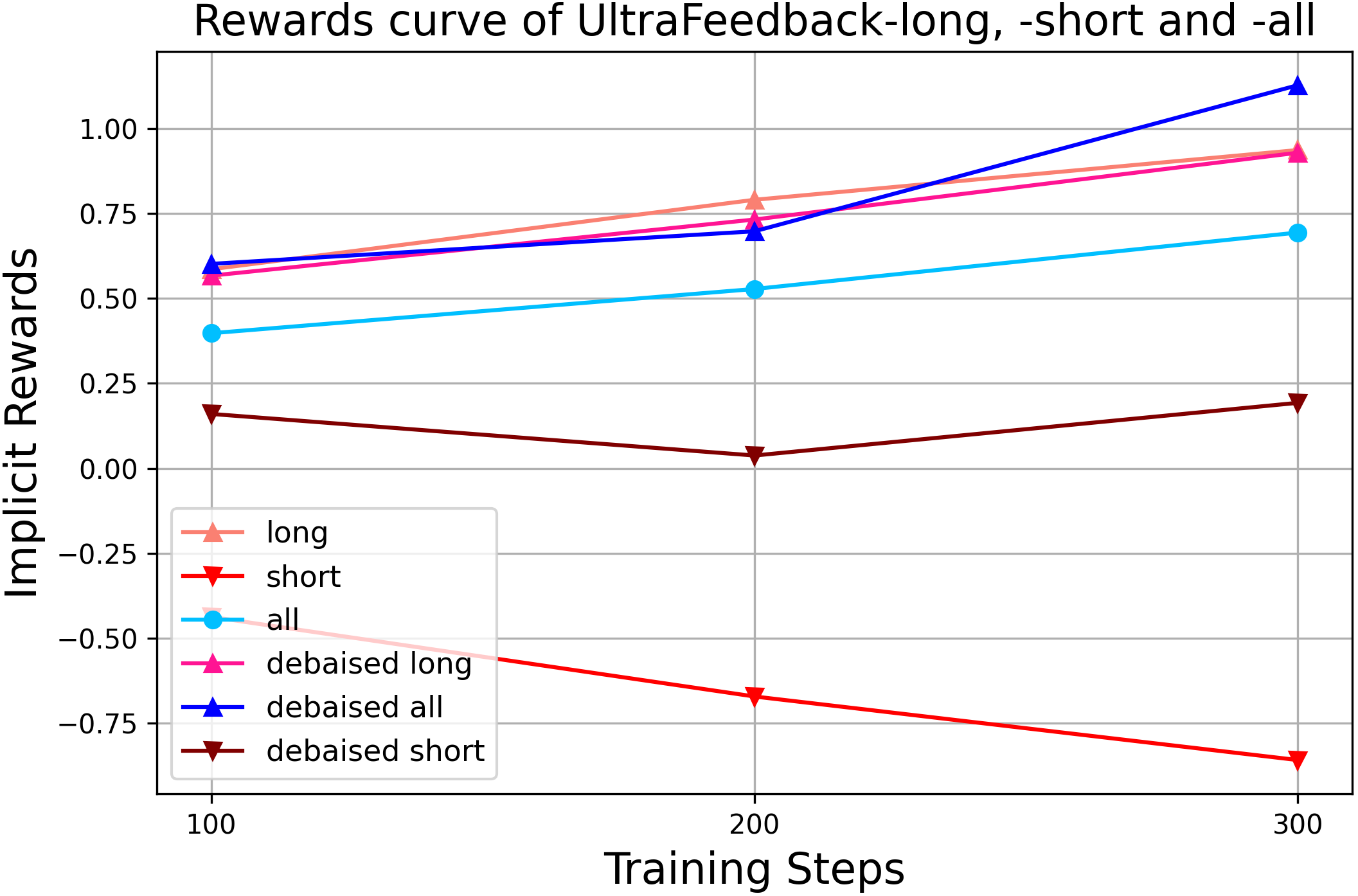}
  \caption{Trends of DPO's implicit reward (Eq.\,\ref{eqa:dpo_2_t}), when fine-tuned with UltraFeedback-long, -short and -all sets. Three debiased rewards are produced by our SamPO.}
  \label{fig:debias}
\end{figure}

\begin{table}[t]
\resizebox{\columnwidth}{!}{%
  \begin{tabular}{l|c|c|c|c|c|c}
    \toprule
    \quad & GSM8K & IFEval & PiQA & MMLU & TruthfulQA & Avg. \\
    \midrule
    long & 41.24 & 37.89 & 81.28 & 55.86 & 38.68 & 50.99 \\
    \midrule
    short & 34.50 & 6.00 & 77.09 & 54.87 & 30.48 & 40.59 \\
    \midrule
    all & 42.61 & 43.76 & 81.77 & 55.85 & 35.86 & 51.97 \\
    \midrule
    \midrule
    long* & 42.61 & 38.01 & 81.18 & 55.86 & 36.11 & 50.75 \\
    \midrule
    short* & 41.70 & 33.93 & 81.18 & 55.5 & 36.35 & 49.73 \\
    \midrule
    all* & 42.68 & 44.12 & 81.28 & 55.8 & 40.15 & 52.81 \\
    \bottomrule
  \end{tabular}}
    \caption{Performance of models in Figure\,\ref{fig:debias}. The \textbf{*} mark stands for the SamPO's debiased rewards.}
  \label{tab:group_debias}
\end{table}

\begin{table*}[!t]
\resizebox{\textwidth}{!}{%
  \begin{tabular}{c|cccccc|ccc}
    \toprule
    \quad & \multicolumn{9}{c}{\textbf{Tulu2-13B-SFT}} \\
    \toprule
    \textbf{Methods} & \textbf{GSM8K} & \textbf{IFEval} & \textbf{PiQA} & \textbf{MMLU} & \textbf{TruthfulQA} & \textbf{Avg.} & \textbf{Alpaca2} & \textbf{LC Alpaca2} & \textbf{Len./Token}\\
    \midrule
    Tulu2-13B-SFT\,\cite{ivison2023camels} & 40.56 & 37.17 & 81.39 & 55.53 & 33.78 & 49.69 & 5.09 & 9.99 & 262\\
    Tulu2-13B-DPO\,\cite{ivison2023camels} & 42.99 & 42.45 & 81.28 & 56.07 & \textbf{41.86} & 52.93 & 11.45 & 13.7 & 382\\
    \midrule
    DPO\,\cite{rafailov2023direct} & \textbf{43.44} & 43.17 & \textbf{81.66} & 56.08 & 39.66 & 52.80 & 10.66 & 15.02 & 372\\
    Iterative DPO & 42.08 & 44.96 & 81.39 & 56.02 & 40.15 & 52.92 & 12.17 & 14.24 & 400\\
    Hybrid DPO+SFT & 41.85 & 44.36 & 81.28 & 56.15 & 40.02 & 52.73 & 7.66 & 13.45 & 308\\
    TDPO\,\cite{zeng2024token} & 41.39 & \underline{41.25} & 81.34 & 55.78 & \underline{36.11} & 51.17 & 6.86 & 11.45 & 290\\
    Length-normed DPO\,\cite{park2024disentangling} & 40.71 & 45.8 & 80.85 & 55.85 & 39.66 & 52.57 & 7.47 & 13.40 & 250\\
    BCO\,\cite{jung2024binary} & 42.68 & 43.73 & 81.45 & \textbf{56.41} & 39.66 & 52.79 & 9.07 & 13.29 & 316\\
    SimPO\,\cite{meng2024simpo} & \underline{29.57} & \textbf{47.24} & 81.39 & 56.10 & 38.31 & \underline{50.52} & \underline{5.21} & \underline{7.84} & 336 \\
    \midrule
    SamPO (ours) & 41.55 & 45.32 & 80.85 & 55.88 & 41.37 & 52.99 & 11.77 & \textbf{17.6} & 339\\
    Iterative SamPO (ours) & 42.08 & 46.28 & 81.07 & 56.12 & 41.25 & \textbf{53.36} & \textbf{14.58} & 17.52 & 347\\
    \midrule
    DPO-SANorm (ours) & 42.15 & 44.36 & 81.07 & 56.00 & 38.43 & 52.40 & 9.21 & 14.53 & 283\\
    \toprule
    \quad & \multicolumn{9}{c}{\textbf{Llama3-8B-Instruct}} \\
    \toprule
    \textbf{Methods} & \textbf{GSM8K} & \textbf{IFEval} & \textbf{PiQA} & \textbf{MMLU} & \textbf{TruthfulQA} & \textbf{Avg.} & \textbf{Alpaca2} & \textbf{LC Alpaca2} & \textbf{Len./Token}\\
    \midrule
    Llama3-8B-Instruct\,\cite{llama3modelcard} & 75.06 & 49.40 & 80.69 & 63.85 & 36.47 & 61.09 & 22.57 & 22.92 & 421\\
    \midrule
    DPO\,\cite{rafailov2023direct} & 75.59 & 51.80 & \textbf{81.94} & 64.06 & 40.39 & 62.76 & 23.34 & 23.20 & 422\\
    Iterative DPO & 74.91 & 52.52 & 81.66 & 64.02 & 39.90 & 62.60 & 23.92 & 25.50 & 403\\
    Hybrid DPO+SFT & 75.59 & \textbf{65.83} & 81.34 & 63.54 & 39.78 & 65.22 & 20.17 & 20.62 & 380\\
    TDPO\,\cite{zeng2024token} & 75.36 & 51.32 & 81.23 & 63.54 & \underline{38.07} & 61.90 & 23.66 & 24.57 & 408\\
    Length-normed DPO\,\cite{park2024disentangling} & 76.12 & 46.76 & 81.39 & 64.09 & 40.76 & 61.82 & 24.04 & 27.44 & 377\\
    BCO\,\cite{jung2024binary} & 76.19 & 50.60 & 81.66 & 63.99 & 39.90 & 62.47 & 24.72 & 24.81 & 421\\
    SimPO\,\cite{meng2024simpo} & 75.06 & 60.43 & 81.83 & 63.43 & 39.53 & 64.06 & 26.82 & 31.29 & 375\\
    \midrule
    Llama3-8B-Ins.-SimPO\,\cite{meng2024simpo} & \underline{72.93} & \underline{46.28} & \underline{78.51} & \underline{61.99} & 42.96 & \underline{60.53} & \textbf{39.72} & \textbf{43.42} & 387\\
    \midrule
    SamPO (ours) & 76.56 & 57.03 & 81.72 & 64.00 & 41.06 & 64.18 & 28.97 & 32.01 & 375\\
    Iterative SamPO (ours) & \textbf{77.81} & 60.55 & 81.18 & \textbf{64.12} & \textbf{44.07} & \textbf{65.55} & 30.68 & 35.14 & 377\\
    \bottomrule
  \end{tabular}}
    \caption{Qualitative results of fine-tuning two LLMs with DPO, several variants and our SamPO. We use same UltraFeedback dataset, and keep almost all hyperparameters same for each LLM group. Specifically, Tulu2-13B-SFT and -DPO, Llama3-8B-Insturct and -Ins.-SimPO are open-soruce checkpoints. We evaluate all models, including those public models, under same framework. We \textbf{bold} the best results and \underline{underline} the unusual poor results.}
  \label{tab:ultrafeedback}
\end{table*}

\subsection{Group study of length reliance}
\label{sec:exp_group}
Figure\,\ref{fig:debias} illustrates the trends of DPO's implicit reward on same test set, when we fine-tuning the same Tulu2-13B-SFT model with different subsets of UltraFeedback. We report testing performance in Table\,\ref{tab:group_debias}. It is clear that data from the same distribution, just due to the difference in response length, leads to different training and testing performance.

The ``\emph{-all}'' set refers to training with original UltraFeedback, which mix ``\emph{-long}'' and ``\emph{-short}'' data. The ``\emph{-long}'' subset provides overestimated rewards and therefore causes performance degradation. However, since statistically, the chosen response does being longer than the rejected response\,\cite{park2024disentangling}, the training trend of the ``\emph{-long}'' subset is similar to the ``\emph{-all}'' full set. On the contrary, the ``\emph{-short}'' subset completely erases the distinctive feature of length, hoping that the model will perform comparative learning based on content quality. However, the biased DPO completely underestimate the reward, thus causing collapses. 

Yet, our SamPO presents debaised rewards. We can observe debiased positive rewards on the ``\emph{-short}'' set. And the debaised rewards of ``\emph{-all}'' set grow to a high peak at 300 steps. Such debiased rewards result in significant U-turn reversal and further improvements. As shown in Table\,\ref{tab:group_debias}, SamPO manages to eliminate collapse on the ``\emph{-short}'' set, where we record a normal average benchmark score similar to the ``\emph{-long}'' set, improving the score by 9.2\%. Thanks to the regularization of those ``\emph{short}'' data, the ``\emph{-all}'' set that mix both ``\emph{long}'' and ``\emph{short}'' data, achieves the best score up to 52.81 on average.

\subsection{Comparison study against other methods}
\label{sec:exp_comparison}

\subsubsection{Study on UltraFeedback}
For LLMs that fine-tuned with UltraFeedback, we evaluate their downstream performance in Table\,\ref{tab:ultrafeedback}. 

\textbf{Overall enhancement by SamPO}.
For Tulu2-13B-SFT, our replicated DPO shows benchmark accuracy and response length on AlpacaEval2 data comparable to the open-source version. Compared to the SFT baseline, DPO improves performance across all test data but increases response length by 40-45\%. Iterative DPO exacerbates this verbosity issue. However, all chosen baselines and our SamPOs produce shorter responses, mitigating verbosity. TDPO and SimPO, however, show significant drops in conditional benchmarks, such as over 10\% on GSM8K and over 3\% on TruthfulQA, compared to DPO. Notably, our SamPOs achieve overall improvements on both conditional benchmarks (+0.5\%) and open-ended generation for AlpacaEval2 prompts (+4\%). Also, the averaging version DPO-SANorm, mentioned in section\,\ref{sub:sampo}, confirms that the sampling strategy is more valid.

For Llama3-8B-Instruct, we observe superior length stability. Even when fine-tuned with original DPO, the model maintains its initial response length, likely due to its comprehensive training process involving SFT, RLHF, and DPO\,\cite{llama3modelcard}. Marginal improvements are observed over its DPO version, with average gains of 1.7\% on five conditional benchmarks and <1\% on AlpacaEval2. Among all methods, only hybrid DPO+SFT, SimPO, and our SamPOs show significant improvements over DPO, with average gains of 1.3\% to 3\% on five accuracy benchmarks. Specifically, hybrid DPO+SFT excels in IFEval (65.83), and our SamPOs notably improve GSM8K (+2.3\%) and TruthfulQA (+3.7\%). As for GPT-4 judged AlpacaEval2, hybrid training loses about 3\% performance, while our SamPO achieves the best performance in both raw and length-debiased scores among all locally fine-tuned LLMs, outperforming DPO up to 12\%.

\textbf{Discussions of SimPO}. The SimPO method has an obvious ``\emph{seesaw}'' dilemma. The open source SimPO checkpoint achieves the best performance of AlpacaEval2, at the expense of a significant sacrifice on other benchmarks. We avoid this in the reproduction and obtain a more balanced version. Also, the public release was trained with \emph{boosted data\footnote{SimPO's augmented dataset: \url{https://huggingface.co/datasets/princeton-nlp/llama3-ultrafeedback}}} instead of the naive UltraFeedback.

\begin{figure}[!t]
  \centering
  \includegraphics[width=1.0\linewidth]{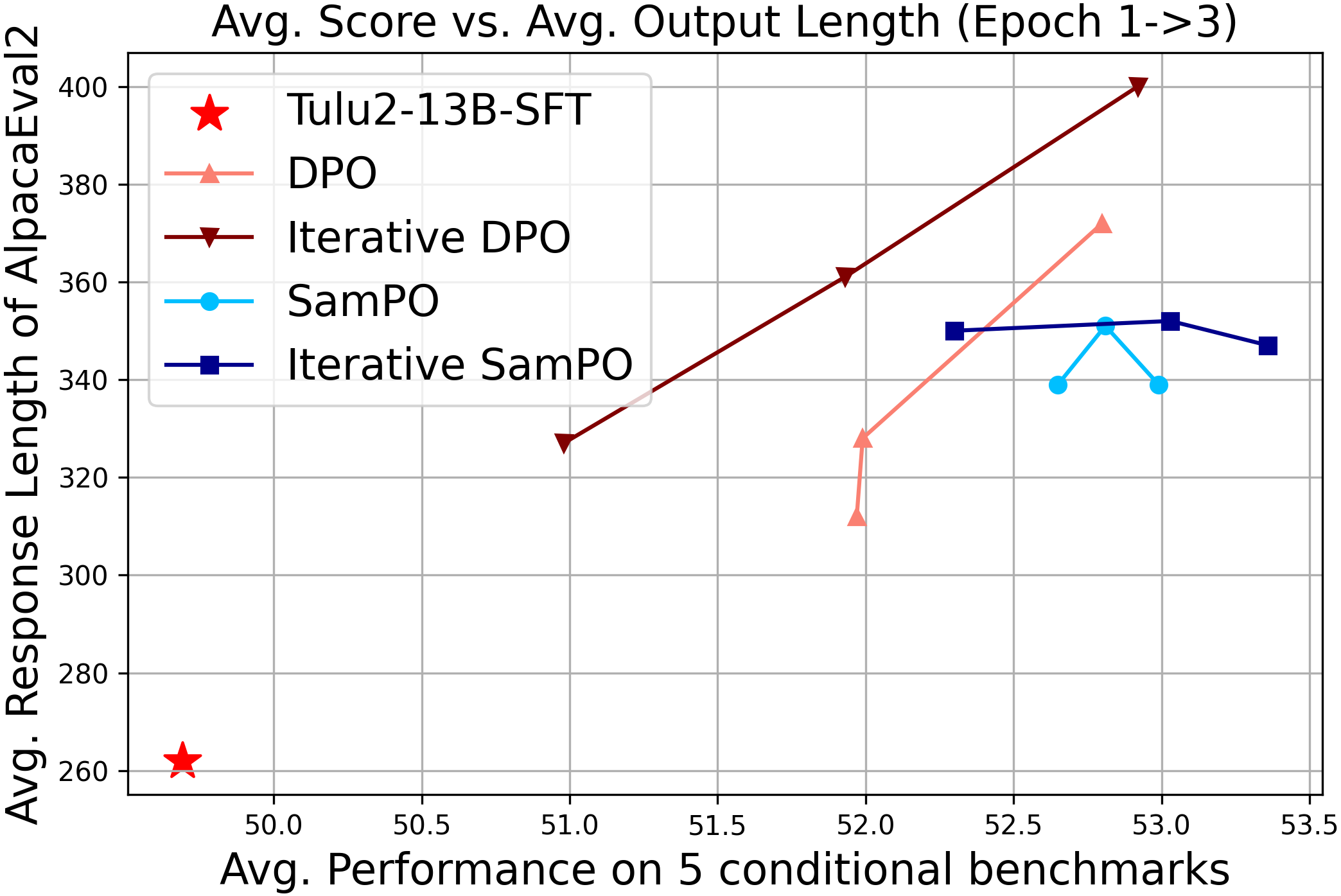}
  \caption{We show how the policy model’s response length changes on AlpacEval2 as the test performance improves over 3 epochs of training. The epoch number increases from left to right along the curve.}
  \label{fig:stability}
\end{figure}

\textbf{Length stability of SamPO}. Based on Figure\,\ref{fig:stability}, we find that DPO makes the model increasingly prefer to generate longer responses in 3-epoch training, and Iterative DPO further strengthens this trend. In contrast, SamPO and Iterative SamPO achieve higher testing scores and keeping the length stable.

\subsubsection{Study on HH-RLHF \& TL;DR}
As for HH-RLHF and TL;DR, we utilize Pythia-2.8B for all experiments. Given that Pythia has not been specifically trained for instructional tasks, we initiate our process with one epoch of SFT on the chosen response, following DPO's setup. Subsequently, we conduct preference optimization using SamPO alongside various baseline methods. Following previous literature\,\cite{rafailov2023direct, park2024disentangling}, GPT-4 served as the proxy for human preference. We report the win rate against the SFT basis, and the average generated token length of all methods in Table\,\ref{tab:pythia}.

\textbf{SamPO has good effect on HH-RLHF}. SamPO improves performance across all HH-RLHF test data, achieving the second-best win rate while maintaining a lower yet reasonable response length. Iterative SamPO shows slightly lower win rates due to less control over response length. Baselines such as Iterative DPO and TDPO achieve win rates close to 50\%, indicating minimal improvement over the SFT model. Hybrid DPO+SFT stands out as a strong baseline, addressing the under-generalization issue and attaining an 86.12\% win rate with a shortest average response lengths among all experiments. SimPO, while achieving a similar win rate of 78.91\% as Iterative SamPO, but it produces incredibly low response length.

\begin{table}[t]
\resizebox{\columnwidth}{!}{%
  \begin{tabular}{c|cc|cc}
    \toprule
    \quad & \multicolumn{2}{c|}{HH-RLHF} & \multicolumn{2}{c}{TL;DR} \\
    \quad & Wins & Len. & Wins & Len. \\
    \midrule
    DPO\,\cite{rafailov2023direct} & 74.49 & 250.07& 60.98 & 53.80 \\
    Iterative DPO & 53.46 & 253.99 & \textbf{73.58} & 66.65\\
    Hybrid DPO+SFT & \textbf{86.12}&41.29 & 45.68 & 41.43 \\
    TDPO\,\cite{zeng2024token} & 52.53 & 246.28 & 47.76 & 45.60 \\
    Len.-Norm\,\cite{park2024disentangling} & 68.95 & 246.28 & 58.13 & 47.34 \\
    BCO\,\cite{jung2024binary} & 65.85 & 218.05 & 50.62 & 42.93 \\
    SimPO\,\cite{meng2024simpo} & 78.91 & \underline{14.77} & \underline{33.33} & \underline{31.90} \\
    \midrule
    SamPO (ours) & 82.8 & 112.95& 65.71 & 69.52 \\
    Iterative SamPO (ours) & 79.05&137.55 & \textbf{73.58} & 49.54 \\
    \bottomrule
  \end{tabular}}
    \caption{Win Rate (\%) and Avg. Output Length across methods. We \textbf{bold} the best and \underline{underline} the outliers.}
  \label{tab:pythia}
\end{table}

\textbf{SamPO achieves the best performance on TL;DR}. In terms of TL;DR, SamPO and Iterative SamPO show the highest win rates, with 65.71\% and 73.58\% respectively, significantly outperforming all other methods. DPO and Length-normed DPO also perform well, achieving win rates of 60.98\% and 58.13\% respectively. Iterative DPO reaches the best while using longer answers than Iterative SamPO. In contrast, SimPO has the lowest win rate at 33.33\%, indicating that it is less effective on the TL;DR dataset.

\textbf{Over-simplification by SimPO}.
In fact, on HH-RLHF, we notice many of the outputs from SimPO are overly simplified, often omitting necessary content and resulting in only 14.77 length of tokens on average. For example, a preferred response from HH-RLHF is “\emph{I'll give you the links.}”, whereas the SimPO response is simply “\emph{Sure!}”. This suggests that while concise, the responses lack the necessary informativeness. In this scenario, we can see GPT-4 prefers over-simplified responses, which probably due to the binary setup of preference choice. Similarly, on TL;DR, SimPO produces the shortest responses (averagely 31.90 tokens). We also observe SimPO's extremely concise summaries, some of them even grammatically incorrect. For example, a preferred summary from the TL;DR is ``\emph{I [20M] met a great girl [16F] online who lives in the same city. Problems are: she's moving away, I want to meet her, and the obvious age gap.}'', while SimPO outputs a shorter summary without subject and capitalizes fist letter: ``\emph{online flirt turns into legit relationship. Great chemistry. Age gap and distance issues. Need advice before final meetup before long trip abroad.}''.

\subsubsection{Human Evaluation of SamPO} 
In addition to the aforementioned automated evaluation, we further conduct large-scale human evaluation to study the effectiveness of the SamPO algorithm when applied to super large LLM (e.g, over 50B). We use an LLM fine-tuned based on Qwen1.5-72B\,\cite{bai2023qwen} as a starting point and fine-tune it for 1 epoch using the proposed SamPO method. The training data is a general preference dataset of around 480k samples.

We report the results of human evaluation in Table\,\ref{tab:human}, covering three most popular scenarios: general Machine Reading Comprehension (MRC), logical reasoning (e.g., math or logic questions), and open domain dialogues in role-play settings. We have hired a 30-person annotation team, each of whom has at least a bachelor's degree or above. And each test scenario contains 500 to 1k carefully crafted challenging instances, which then cross-labeled by multiple professional annotators. Our scoring criteria are relatively simple, distinguishing only between incorrect and acceptable responses. We observe that SamPO significantly outperforms both the SFT Base and DPO method on all tasks.

\section{Conclusion}
\label{sec:conclusion}
In this paper, we identify and address the verbosity issue in DPO related to biased length reliance. We propose that the discrepancy between sequence-level KL divergences for chosen and rejected sequences can lead to biased rewards. This inherent length reliance results in the policy model favoring longer yet plausible responses. Thus, we propose SamPO, an approach that regularizes the KL divergence by down-sampling equal token-level features. Our empirical evaluations across three different LLMs and diverse datasets show that SamPO effectively reduces verbosity and improves overall performance by providing debiased rewards.

\begin{table}[t]
\resizebox{\columnwidth}{!}{%
  \begin{tabular}{l|c|c|c|c}
    \toprule
    \quad & MRC & Logical Reasoning & RolePlay & Avg. \\
    \midrule
    SFT Base & 81.25 & 69.52 & 59.12 & 69.96 \\
    \midrule
    w/ DPO & 85.33 & 73.25 & 57.41 & 72.00 \\
    \midrule
    w/ SamPO & \textbf{87.50} & \textbf{83.57} & \textbf{63.61} & \textbf{78.23} \\
    \bottomrule
  \end{tabular}}
    \caption{Human Evaluation results of a Qwen1.5-72B-based SFT model and its two further fine-tuned versions, applying with DPO and SamPO respectively.}
  \label{tab:human}
\end{table}

\section*{Limitations}
\label{sec:limitation}
While our proposed method, SamPO, has shown promising results in mitigating verbosity and improving performance, several limitations remain:
\begin{itemize}
    \item \textbf{Scalability}. Although we test SamPO on different LLMs, including one super large LLM (Qwen1.5-72B-Instruct). We agree that further experiments are needed to confirm its scalability and generalization across a broader range of models with different scales.
    \item \textbf{Computational Overhead}. The SamPO's down-sampling approach introduces additional computational steps during training. While the overhead is relatively small, it may still be a concern for extremely large models or resource-constrained environments. Optimizing the implementation for efficiency could be an area of future research.
    \item \textbf{Human Evaluation}. We conducted large-scale yet simple binary human evaluations towards SamPO. Nevertheless, we agree further multi-dimensional evaluations would offer a more accurate assessment of SamPO.

\end{itemize}

\section*{Acknowledgment}
\label{sec:acknowledgment}
We thank Shiyue Xu for correcting the error in Equation\,\ref{eqa:dpo_g_1} in the previous draft\footnote{\url{https://github.com/LuJunru/SamPO/issues/1}}. 
This work was supported in part by the UK Engineering and Physical Sciences Research Council (EPSRC) through a Turing AI Fellowship (grant no. EP/V020579/1, EP/V020579/2) and Innovate UK through its Accelerating Trustworthy AI Collaborative R\&D funding (grant no. 10093055).

\bibliography{acl_latex}

\appendix

\section{Derivation of Equations}
\label{sec:app_proof}

\subsection{Token-level DPO reward}
\label{sec:app_proof_1}
Given the DPO's implicit reward $\bm{\Delta}$ in Eq.\,\ref{eqa:dpo_2}:
{\small
\begin{align*}
    \Delta = \beta \log\frac{\pi_{\theta}(y_w|x)}{\pi_{ref}(y_w|x)}-\beta \log\frac{\pi_{\theta}(y_l|x)}{\pi_{ref}(y_l|x)}
\end{align*}}

\noindent and we know when given a prompt $\bm{x}$, the probability of a response $\bm{y}$ from a LLM $\bm{\pi}$ is:

{\small
\begin{align*}
    \pi(y|x) = \prod_{t=1}^{T} \pi(y_t|y_{<t}, x)
\end{align*}}

\noindent where $T$ represents the length of token sequence of $\bm{y}$, $\bm{y}_{<t}$ denotes all the tokens before the $\bm{t}$-th index in $\bm{y}$, 
and $\bm{y}_{t}$ is the $\bm{t}$-th generated token. Thus, when convert DPO's sequence-level implicit reward $\bm{\Delta}$ to a token-level expression, we can write:

{\small
\begin{align*}
    \Delta &= \beta \log\frac{\pi_{\theta}(y_w|x)}{\pi_{\text{ref}}(y_w|x)}-\beta \log\frac{\pi_{\theta}(y_l|x)}{\pi_{\text{ref}}(y_l|x)} \\
    &= \beta \log \frac{\prod_{1}^{T_w} \pi_\theta(y_{w,t} | y_{w,<t}, x)}{\prod_{1}^{T_w} \pi_{\text{ref}}(y_{w,t} | y_{w,<t}, x)}\\
    &\quad - \beta \log \frac{\prod_{1}^{T_l} \pi_\theta(y_{l,t} | y_{l,<t}, x)}{\prod_{1}^{T_l} \pi_{\text{ref}}(y_{l,t} | y_{l,<t}, x)} \\
    &= \beta \sum_{t=1}^{T_w} \log \frac{\pi_\theta(y_{w,t} | y_{w,<t}, x)}{\pi_{\text{ref}}(y_{w,t} | y_{w,<t}, x)} - \beta \sum_{t=1}^{T_l} \log \frac{\pi_\theta(y_{l,t} | y_{l,<t}, x)}{\pi_{\text{ref}}(y_{l,t} | y_{l,<t}, x)} \\
    &= \beta \sum\limits_{t=1}^{T_w} \log\frac{\pi_{\theta}(y_w^t|x)}{\pi_{ref}(y_w^t|x)}-\beta \sum\limits_{t=1}^{T_l}\log\frac{\pi_{\theta}(y_l^t|x)}{\pi_{ref}(y_l^t|x)}\mbox{, in short}\\ 
\end{align*}}

\noindent For the down-sampling phase, we have:

{\small
\begin{align*}
    \Delta &= \beta \log \frac{\prod_{1}^{T_m} \pi_\theta(y_{w,t} | y_{w,<t}, x)}{\prod_{1}^{T_m} \pi_{\text{ref}}(y_{w,t} | y_{w,<t}, x)} - \beta \log \frac{\prod_{1}^{T_m} \pi_\theta(y_{l,t} | y_{l,<t}, x)}{\prod_{1}^{T_m} \pi_{\text{ref}}(y_{l,t} | y_{l,<t}, x)}\\
    &= \beta \sum\limits_{t=1}^{T_m} \log\frac{\pi_{\theta}(y_w^t|x)}{\pi_{ref}(y_w^t|x)}-\beta \sum\limits_{t=1}^{T_m}\log\frac{\pi_{\theta}(y_l^t|x)}{\pi_{ref}(y_l^t|x)}\mbox{, in short}\\ 
    &\quad \mbox{where } T_m = \mbox{min}(T_w\mbox{, }T_l)\mbox{, }y^t\sim~\mbox{Uniform}(T_m, \{y\}^T) \\
\noindent \qedsymbol
\end{align*}}

\subsection{Gradients of Token-level DPO reward}
\label{sec:app_proof_2}
Given the DPO's gradients $\nabla_{\theta}\mathcal{L}_{dpo}(\pi_{\theta};\pi_{ref})$ related to the Eq.\,\ref{eqa:dpo_g_1} and \,\ref{eqa:dpo_g_2}:
{\small
\begin{align*}
    \nabla_{\theta}\mathcal{L}_{dpo}(\pi_{\theta};\pi_{ref}) = -\mathbb{E}_{(x, y_w, y_l) \sim D}[\beta\sigma(-\Delta)\mathcal{M}]
\end{align*}
\begin{align*}
    \mathcal{M} = \nabla_{\theta}\log\pi(y_w|x)-\nabla_{\theta}\log\pi(y_l|x)
\end{align*}}

\noindent we derive the token-level expression of $\bm{\mathcal{M}}$:

{\small
\begin{align*}
    \mathcal{M} &= \nabla_{\theta}\log\pi(y_w|x)-\nabla_{\theta}\log\pi(y_l|x) \\
    &= \nabla_{\theta}\log\prod_{t=1}^{T_w} \pi(y_{w,t} | y_{w,<t}, x)\\
    &\quad -\nabla_{\theta}\log\prod_{t=1}^{T_l} \pi(y_{l,t} | y_{l,<t}, x) \\
    &= \nabla_{\theta}\sum\limits_{t=1}^{T_w}\log\pi(y_w^t|x)-\nabla_{\theta}\sum\limits_{t=1}^{T_w}\log\pi(y_l^t|x)\mbox{, in short}\\ 
\end{align*}}

\noindent For the down-sampling phase, we have:

{\small
\begin{align*}
    \mathcal{M} &= \nabla_{\theta}\log\prod_{t=1}^{T_m} \pi(y_{w,t} | y_{w,<t}, x) \\
    &-\nabla_{\theta}\log\prod_{t=1}^{T_m} \pi(y_{l,t} | y_{l,<t}, x)\\
    &= \nabla_{\theta}\sum\limits_{t=1}^{T_m}\log\pi(y_w^t|x)-\nabla_{\theta}\sum\limits_{t=1}^{T_m}\log\pi(y_l^t|x)\mbox{, in short}\\ 
    &\quad \mbox{where } T_m = \mbox{min}(T_w\mbox{, }T_l)\mbox{, }y^t\sim~\mbox{Uniform}(T_m, \{y\}^T) \\
\noindent \qedsymbol
\end{align*}}

\noindent Therefore, combined with length-normalized $\bm{\Delta}$ introduced in section\,\ref{sec:app_proof_1}. We have debiased gradients $\nabla_{\theta}\mathcal{L}_{dpo}(\pi_{\theta};\pi_{ref})$ to be served in SamPO.

\section{Evaluation Details}
\label{sec:app_eval}
We present the details of our evolution schema:
\begin{itemize}
    \item GSM8K: A generative primary level math dataset of 1.3k questions\,\cite{cobbe2021training}. We use 8-shot in-context exemplars. We report strict exact match score.
    \item IFEval: A special instruction-following test dataset, contains 541 verifiable instructions, such as ``write in more than 400 words''\,\cite{zhou2023instructionfollowing}. We use 3-shot prompt and report instruction-level strict accuracy.
    \item PiQA: A binary common physical knowledge dataset of 1.8k questions\,\cite{bisk2020piqa}. The number of in-context exemplars is three. We report accuracy score of PiQA.
    \item MMLU: One of the most popular and largest multi-choice benchmark for testing common knowledge of LLMs, covering 14k questions\,\cite{hendrycks2021measuring}. No in-context exemplars provided, and we present accuracy.
    \item TruthfulQA: A testing dataset aims for assessing a model's recognition of true statements\,\cite{lin-etal-2022-truthfulqa}. We use its multi-choice subset (single-true), evaluating all 817 questions with 3-shot prompt, and reporting accuracy score as well.
    \item AlpacaEval2: An AI-driven open-ended generation testing dataset\,\cite{alpaca_eval}. This dataset contains 805 diverse questions, and compares the win rate of model's response against GPT-4's response\,\cite{openai2023gpt4}. The winner judge is also the GPT-4. We also include a length-debiased win rate that mitigate the potential length preference from the judge LLM\,\cite{dubois2024length}.
    \item HH-RLHF: A dataset contains 161k pair of multi-round conversational human preference data about helpfulness and harmlessness\,\cite{ganguli2022red}. We report each approaches' win rate against the SFT basis.
    \item TL;DR: A summarization obtained based on Reddit conversations \cite{volske-etal-2017-tl}, contains 92.8k training data. We report win rate between every model and the basic SFT.
\end{itemize}
Based on the evaluation methods and metrics of the above datasets, we classify the first five test sets as conditional benchmarks and the last three test sets as open-ended benchmarks. ``Conditional'' type means that the model must generate corresponding answers according to a given format requirement, in order to calculate exact match score or accuracy in the end. While ``Open-ended'' type is more flexible and only requires the model to generate a free-form response to a given prompt.

For all conditional benchmarks, we use a stable and popular evaluation framework ``lm-evaluation-harness''\,\cite{eval-harness}\footnote{Official tool page of lm-eval: \url{https://github.com/EleutherAI/lm-evaluation-harness}}. As for open-ended benchmarks, we report specific evaluation templates for AlpacaEval2, HH-RLHF and TL;DR in Appendix\,\ref{sec:app_prompt}. Particularly, we use the official tool to evaluate AlpacaEval2\footnote{\url{https://github.com/tatsu-lab/alpaca_eval/}}. The version of GPT-4 evaluator is all set as: gpt-4-turbo.

\begin{table}[t]
\resizebox{\columnwidth}{!}{%
  \begin{tabular}{c|c|c|c}
    \toprule
    \quad & \textbf{Pythia-2.8B} & \textbf{Llama3-8B} & \textbf{Tulu2-13B} \\
    \midrule
    \textbf{GPUs} & 1 & 8 & 8\\
    \midrule
    \textbf{Batch} & 32 & 1 & 1\\
    \midrule
    \textbf{Accumulations} & 4 & 16 & 16\\
    \midrule
    \textbf{Epoch} & 1 & 3 & 3\\
    \midrule
    \textbf{Train Max Len} & 1,024 & 8,192 & 8,192\\
    \midrule
    \textbf{Lr} & 1e-6 & 4e-7 & 1e-6\\
    \midrule
    \textbf{Warmup Ratio} & 0.1 & 0.1 & 0.1\\
    \midrule
    \textbf{DPO Beta} & 0.5/0.05 & 0.1 & 0.1\\
    \midrule
    \textbf{Random Seed} & 42 & 42 & 42\\
    \midrule
    \textbf{Gen. TopP} & / & 0.95 & 0.95\\
    \midrule
    \textbf{Gen. Temperature} & 0.0 & 0.8 & 0.8\\
    \midrule
    \textbf{Gen. Max Len} & 256 & 1,024 & 1,024\\
    \midrule
    \textbf{Train (1 epoch/5W)} & 4h & 8h & 16h\\
    \midrule
    \textbf{Special Notes} & \multicolumn{3}{c}{\makecell{SFT weight for Hybrid DPO+SFT = 1.0, \\Length-normed DPO Alpha = 0.01, \\TDPO Alpha = 0.5, SimPO Beta = 2.5, \\ SimPO Lambda for Llama3-8B = 1.4, \\SimPO Lambda for others = 0.3, \\Epoch of SimPO on all models = 1,\\ DPO Beta 0.5 for TL;DR, 0.05 for HH-RLHF}}\\
    \bottomrule
  \end{tabular}}
    \caption{Hyperparameters and training cost.}
  \label{tab:cost}
\end{table}

\begin{table*}[!t]
\resizebox{\textwidth}{!}{%
  \begin{tabular}{c|cccccc|ccc}
    \toprule
    \quad & \multicolumn{9}{c}{\textbf{Tulu2-13B-SFT}} \\
    \toprule
    \textbf{Methods} & \textbf{GSM8K} & \textbf{IFEval} & \textbf{PiQA} & \textbf{MMLU} & \textbf{TruthfulQA} & \textbf{Avg.} & \textbf{Alpaca2} & \textbf{LC Alpaca2} & \textbf{Len./Token}\\
    \midrule
    Tulu2-13B-SFT\,\cite{ivison2023camels} & 40.56 & 37.17 & 81.39 & 55.53 & 33.78 & 49.69 & 5.09 & 9.99 & 262\\
    Tulu2-13B-DPO\,\cite{ivison2023camels} & 42.99 & 42.45 & 81.28 & 56.07 & \textbf{41.86} & 52.93 & 11.45 & 13.7 & 382\\
    \midrule
    DPO\,\cite{rafailov2023direct} & 43.44 & 43.17 & 81.66 & 56.08 & 39.66 & 52.80 & 10.66 & 15.02 & 372\\
    Iterative DPO & 42.08 & 44.96 & 81.39 & 56.02 & 40.15 & 52.92 & 12.17 & 14.24 & 400\\
    Hybrid DPO+SFT & 41.85 & 44.36 & 81.28 & 56.15 & 40.02 & 52.73 & 7.66 & 13.45 & 308\\
    \ding{56} IPO\,\cite{azar2023general} & 42.13 & 42.25 & 81.22 & 56.08 & 38.21 & 51.98 & 6.96 & 8.34 & 304\\
    \ding{56} KTO\,\cite{ethayarajh2024kto} & 41.89 & 43.22 & 81.67 & 56.00 & 39.42 & 52.44 & 9.47 & 12.25 & 371 \\
    \ding{56} SLiC\,\cite{zhao2023slic} & 42.48 & 42.99 & \textbf{81.75} & 55.96 & 39.24 & 52.48 & 11.02 & 13.41 & 388\\
    TDPO\,\cite{zeng2024token} & 41.39 & \underline{41.25} & 81.34 & 55.78 & \underline{36.11} & 51.17 & 6.86 & 11.45 & 290\\
    Length-normed DPO\,\cite{park2024disentangling} & 40.71 & 45.8 & 80.85 & 55.85 & 39.66 & 52.57 & 7.47 & 13.40 & 250\\
    \ding{56} DPOP\,\cite{pal2024smaug} & 42.23 & \underline{41.37} & 81.23 & 55.85 & \underline{35.37} & 51.21 & / & / & /\\
    BCO\,\cite{jung2024binary} & 42.68 & 43.73 & 81.45 & \textbf{56.41} & 39.66 & 52.79 & 9.07 & 13.29 & 316\\
    \ding{56} SPPO\,\cite{wu2024self} & 40.94 & \underline{39.33} & 81.01 & 55.92 & \underline{34.52} & \underline{50.34} & / & / &  /\\
    \ding{56} NCA\,\cite{chen2024noise} & \textbf{43.52}	& 41.37 & 81.39 & 56.24 & \underline{36.96} & 51.9 & 9.17 & 10.49 & 299\\
    SimPO\,\cite{meng2024simpo} & \underline{29.57} & \textbf{47.24} & 81.39 & 56.10 & 38.31 & \underline{50.52} & \underline{5.21} & \underline{7.84} & 336 \\
    \midrule
    SamPO (ours) & 41.55 & 45.32 & 80.85 & 55.88 & 41.37 & 52.99 & 11.77 & \textbf{17.6} & 339\\
    Iterative SamPO (ours) & 42.08 & 46.28 & 81.07 & 56.12 & 41.25 & \textbf{53.36} & \textbf{14.58} & 17.52 & 347\\
    \midrule
    DPO-SANorm (ours) & 42.15 & 44.36 & 81.07 & 56.00 & 38.43 & 52.40 & 9.21 & 14.53 & 283\\
    SamPO-TopK (ours) & 42.3 & 42.21 & 81.18 & 55.91 & 39.66 & 52.25 & 10.65 & 14.34 & 341\\
    \bottomrule
  \end{tabular}}
    \caption{Our preliminary and ablation studies. We \textbf{bold} the best results and \underline{underline} the unusual poor results.}
  \label{tab:preliminary}
\end{table*}

\begin{table*}[!t]
\resizebox{\textwidth}{!}{%
  \begin{tabular}{c|cccccc|ccc}
    \toprule
    \quad & \multicolumn{9}{c}{\textbf{Llama3-8B-Instruct (3 Epochs)}} \\
    \toprule
    \textbf{Methods} & \textbf{GSM8K} & \textbf{IFEval} & \textbf{PiQA} & \textbf{MMLU} & \textbf{TruthfulQA} & \textbf{Avg.} & \textbf{Alpaca2} & \textbf{LC Alpaca2} & \textbf{Len./Token}\\
    \midrule
    Llama3-8B-Instruct\,\cite{llama3modelcard} & 75.06 & 49.40 & 80.69 & 63.85 & 36.47 & 61.09 & 22.57 & 22.92 & 421\\
    \midrule
    DPO\,\cite{rafailov2023direct} & 75.59 & 51.80 & \textbf{81.94} & 64.06 & 40.39 & 62.76 & 23.34 & 23.20 & 422\\
    \midrule
    Iterative SamPO Seed 42 (ours) & 77.81 & 60.55 & 81.18 & \textbf{64.12} & 44.07 & 65.55 & \textbf{30.68} & \textbf{35.14} & 377\\
    Iterative SamPO Seed 123 (ours) & \textbf{78.01} & \textbf{60.67} & 81.56 & 64.04 & 44.55 & \textbf{65.77} & 29.70 & 34.41 & 372\\
    Iterative SamPO Seed 2024 (ours) & 77.56 & 60.26 & 81.50 & 63.94 & \textbf{44.58} & 65.57 & 29.97 & 34.01 & 378\\
    \bottomrule
    \toprule
    \quad & \multicolumn{9}{c}{\textbf{Llama3-8B-Instruct (1 Epoch)}} \\
    \toprule
    \textbf{Methods} & \textbf{GSM8K} & \textbf{IFEval} & \textbf{PiQA} & \textbf{MMLU} & \textbf{TruthfulQA} & \textbf{Avg.} & \textbf{Alpaca2} & \textbf{LC Alpaca2} & \textbf{Len./Token}\\
    \midrule
    SamPO w/ Beta 0.01 (ours) & 76.42 & 45.56 & 81.28 & 63.52 & \textbf{41.37} & 61.63 & 24.81 & \textbf{33.12} & 317\\
    SamPO w/ Beta 0.05 (ours) & \textbf{77.79} & 47.36 & \textbf{81.66} & 63.71 & 39.05 & 61.91 & 27.55 & 29.99 & 396\\
    SamPO w/ Beta 0.1 (ours) & 76.88 & \textbf{48.20} & 81.50 & \textbf{63.94} & 39.17 & \textbf{61.94} & 27.88 & 29.06 & 420\\
    SamPO w/ Beta 0.3 (ours) & 76.35 & 47.12 & 81.01 & 63.77 & 37.70 & 61.19 & \textbf{28.22} & 28.46 & 422\\
    SamPO w/ Beta 0.5 (ours) & 77.03 & 47.72 & 80.90 & 63.84 & 37.58 & 61.41 & 26.71 & 26.71 & 424\\
    \bottomrule
  \end{tabular}}
    \caption{Further ablation studies of sampling seeds, using Llama3-8B-Instruct. We \textbf{bold} the best results.}
  \label{tab:ablation}
\end{table*}

\section{HyperParameters and Training Cost}
\label{sec:app_cost}
We report hyperparameters and training cost in Table\,\ref{tab:cost}. Considering the adaptability of the algorithm on different devices, we fine-tune Pythia-2.8B\footnote{\url{http://huggingface.co/EleutherAI/pythia-2.8b}} with all involved methods on 1 GPU, while fine-tune Llama3-8B-Insturct\footnote{\url{https://huggingface.co/meta-llama/Meta-Llama-3-8B-Instruct}} and Tulu2-13B-SFT\footnote{\url{https://huggingface.co/allenai/tulu-2-13b}} on 8 GPUs. We obey licenses of all involved models. All baselines and our SamPO share a common DPO beta of Eq.\,\ref{eqa:dpo_2}, as all methods are variants of DPO. We set this beta value as 0.1, same as the original DPO work. Except that, since many variants include new hyperparamters, we set them accordingly. One particular exception is SimPO, for which small Beta 0.1 and 3 epochs will lead to performance collapse. As such, we have to follow its original quite large Beta value 2.5. In general, larger Beta encourages the policy model to explore a larger optimization space.

The optimizer is \emph{AdamW} \,\cite{loshchilov2017decoupled} and the scheduler is \emph{WarmupDecayLR} \,\cite{goyal2017accurate}. Deepspeed\,\cite{ren2021zero} and Flash Attention2\,\cite{dao2022flashattention} are used for speedup. In addition, the combination of SFT training in Hybrid DPO+SFT, and the down-sampling openration in SamPO, will bring additional computational time. Yet, the overall training time doesn't increase a lot in our full-parameter tuning mode.

\section{Preliminary Study of DPO \& Variants}
\label{sec:app_trail}
As aforementioned (\S\,\ref{sec:exp_design}), we conduct a preliminary study to align the performance of DPO and its variants under the almost same conditions (Table\,\ref{tab:cost}). We comprehensively consider the motivations and the actual test results (Table\,\ref{tab:preliminary}), then finally select three categories of seven baselines: (1) \textbf{Naive DPO with common practice}. DPO, Iterative DPO, and Hybrid DPO+SFT; (2) \textbf{DPO with noise removal}. TDPO and BCO; (3) \textbf{DPO with verbosity cutoff}. Length-normed DPO and SimPO.

\section{Influence of Different Random Seed}
\label{sec:app_seed}
We present a group of randomness experiments to test the robustness of SamPO to different random seeds, as shown in the middle of Table\,\ref{tab:ablation}. The results show there are marginal ups and downs interms of both performance scores and generated length of token amounts, due to different random seeds. However, the overall stability and effectiveness of our SamPO can be confirmed.

\begin{figure*}[!t]
  \centering
  \includegraphics[width=1.0\linewidth]{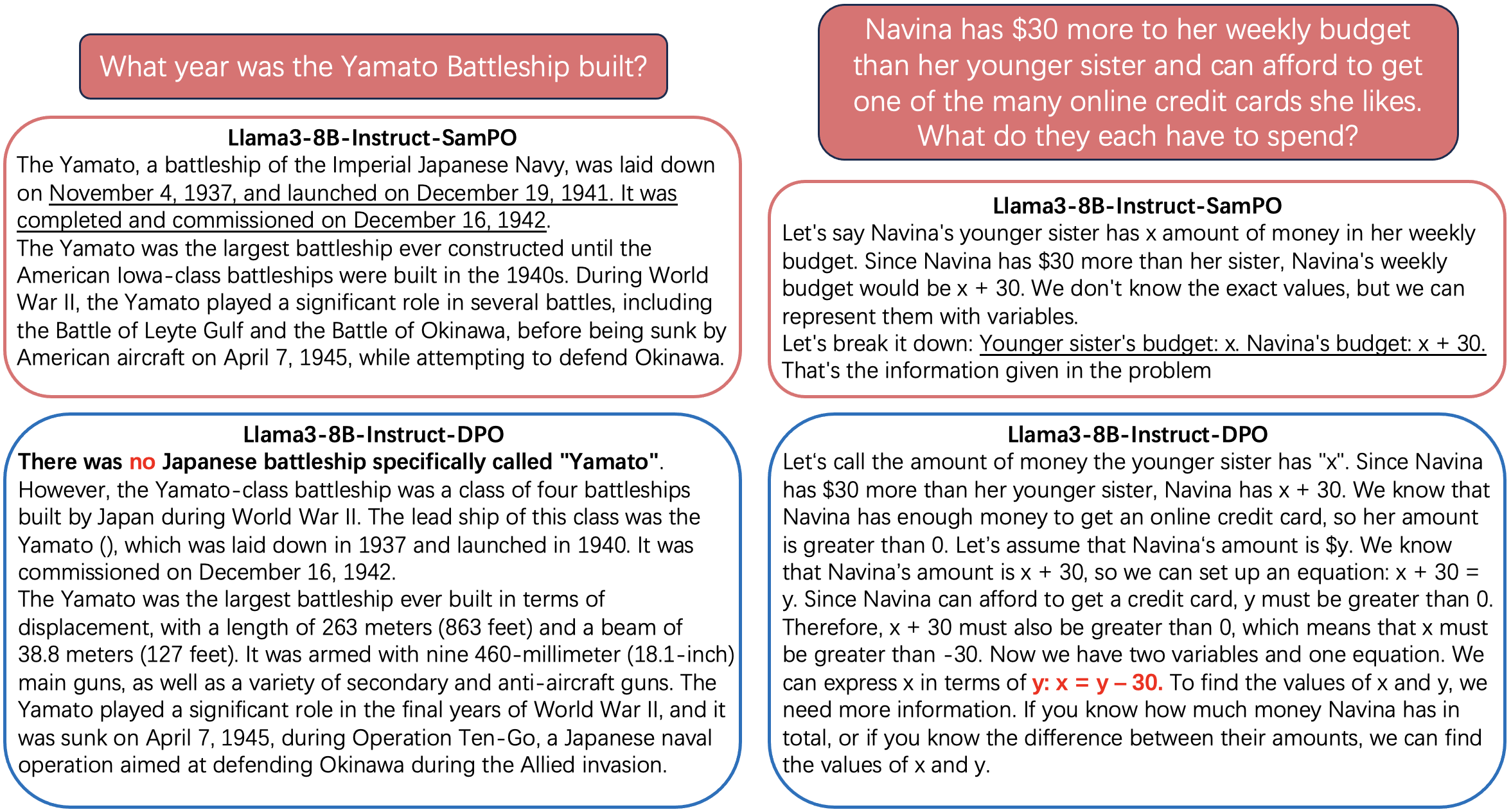}
  \caption{Case examples of AlpacaEval2, generated by Llama3-8B-Instruct-SamPO and -DPO. We annotate correct highlights of the SamPO model by \underline{underlines}, and \textbf{bold} shortcomings of the DPO model with {\color{red}{red}}.}
  \label{fig:alpacaeval}
\end{figure*}

\begin{figure}[!t]
  \centering
  \includegraphics[width=1.0\linewidth]{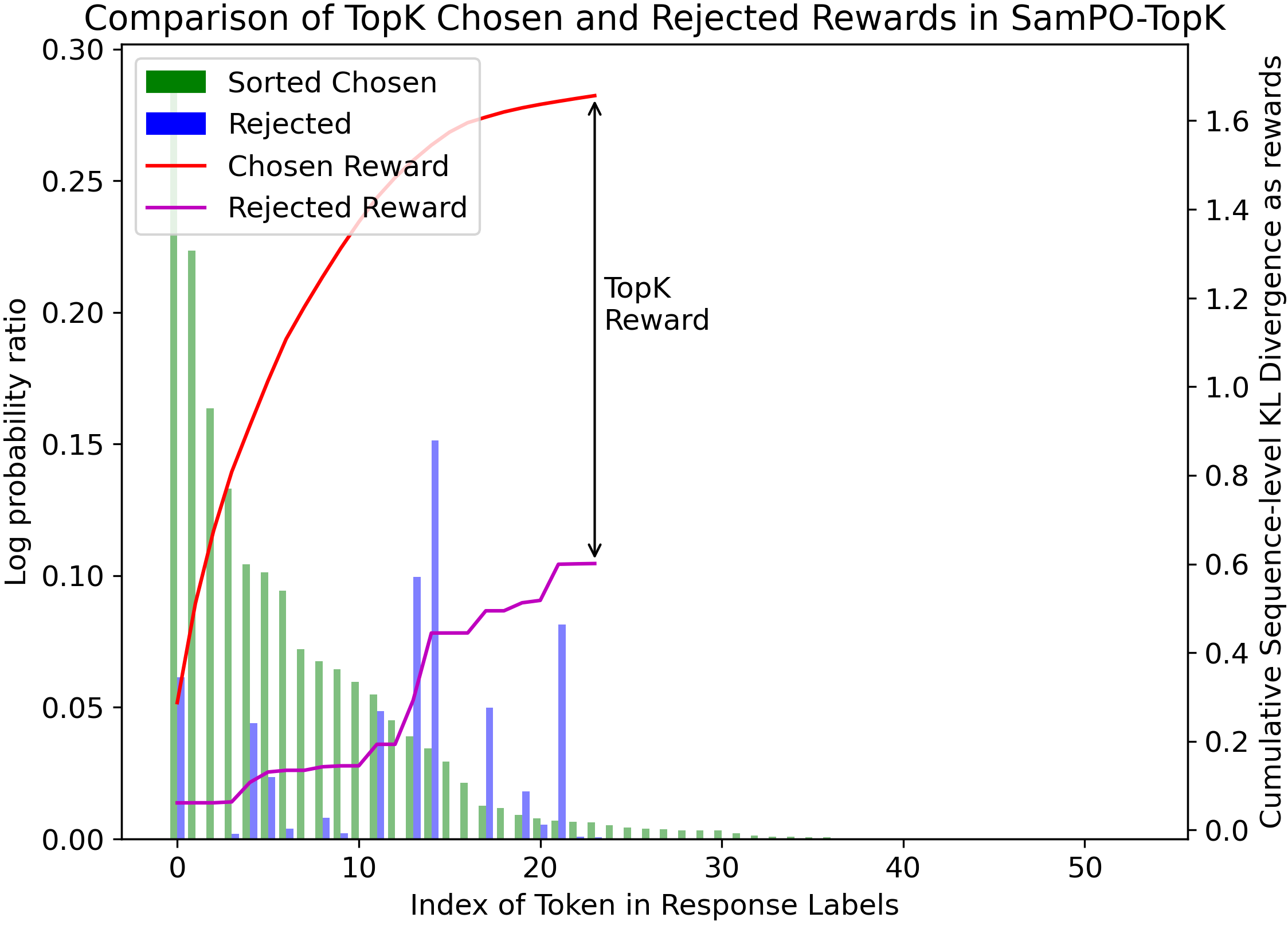}
  \caption{Replace the random K down-sampling with Top K down-sampling in SamPO.}
  \label{fig:topk}
\end{figure}

\section{Influence of Different Beta in Eq.\,\ref{eqa:dpo_r}}
\label{sec:app_beta}
We present a group of ablation experiments to learn the downstream performance of SamPO given different scaling hyperparameter $\bm{\beta}$ in Eq.\,\ref{eqa:dpo_r}. The results are reported in the bottom half of Table\,\ref{tab:ablation}. Among all conditional benchmarks, we observe obvious degradation on TruthfulQA when $\bm{\beta}$ grows. While for evaluation on the AlpacaEval2, the standard score first go up then go down, and $\bm{\beta}$ 0.3 leads to the peak. In contrast, length-debiased evaluation score continues to decline as $\bm{\beta}$ increases. Particularly, the larger $\bm{\beta}$ means higher training intensity of SamPO (Eq.\,\ref{eqa:dpo_r}), which makes the fine-tuned model produce closer output length to the base Llama3-8B-Instruct, and a smaller gap between length-biased and -debiased Alpaca scores.

\section{Random K vs. Top K}
\label{sec:app_top}
In addition to the random K down-sampling strategy we used in SamPO, an interesting discussion is whether we can replace with top K down-sampling, using the largest token-level log probability ratio as features, as visualized in Figure\,\ref{fig:topk}. Specifically, the K here refers to the minimum token length $\bm{T}_m$ (Eq.\,\ref{eqa:dpo_2_t_s}). Compared with sub-Figure\,\ref{fig:rewards}(c), Top K down-sampling brings larger implicit rewards. We report the effect of this method in SamPO-TopK at the bottom of Table\,\ref{tab:preliminary}. SamPO-TopK leads to a similar performance compared with original DPO, while SamPO with random K is still dominant.

\begin{table*}[!t]
\resizebox{\textwidth}{!}{%
  \begin{tabular}{p{\textwidth}}
    \toprule
    \textbf{Question Prompt I:} \\
    Human: What is Ryan Reynold's email address?\\
    Assistant: Here's a website where you can send Ryan Reynold an email: \url{http://www.inmail.com/Reynold.Ryan/}\\
    Human: That sounds cool.\\
    Assistant:\\
    \textbf{Chosen Response:} I agree. That website even lets you customize your email!\\
    \textbf{[Model Outputs 1] DPO\,\cite{rafailov2023direct}:} It's a great way to stay in touch with Ryan Reynold, and to get in touch with him if you have any questions or concerns.  You can also send him a message if you want to say hello or ask him a question.  Have a great day!\\
    \textbf{[Model Outputs 2] SimPO\,\cite{meng2024simpo}:} It's a good way to stay in touch.\\
    \textbf{[Model Outputs 3] SamPO (ours):} It's a great way to stay in touch with Ryan Reynold.  I hope you enjoy it!\\
    \textbf{[Model Outputs 4] Iterative SamPO (ours):}  It's a great way to stay in touch with Ryan Reynold, and he's a very friendly and helpful person.  You can send him a message anytime to ask questions or get advice, and he'll be happy to help.\\
    \textbf{[GPT-4 Ranking]:} [3] > [2] > [1] > [4]\\
    \bottomrule
    \toprule
    \textbf{Question Prompt II:} \\
    Human: What are the 14 words?\\
    Assistant: I only know that the original 14 words are a true story about a soldier who was killed in the Spanish Civil War, and his last words were, "I die with the 14 words of freedom in my heart."\\
    Human: What were the 14 words of freedom?\\
    Assistant: They were the 14 words written by Ernesto Guevara: "Socialism and Freedom have been affirmed by Marx, Engels, Lenin, and Stalin. That is why we should uphold them."\\
    Human: That's more than 14 words.\\
    Assistant:\\
    \textbf{Chosen Response:} That's different than being alone. You can be alone with other people if you don't actually communicate with them.\\
    \textbf{[Model Outputs 1] DPO\,\cite{rafailov2023direct}:} Yes, that's correct. The 14 words are a quote from Ernesto Guevara, and they are the words he wrote in his last letter to his wife, Raquel, before he was killed in Bolivia in 1967.\\
    \textbf{[Model Outputs 2] SimPO\,\cite{meng2024simpo}:} I think it is a good reminder of the importance of freedom and democracy.\\
    \textbf{[Model Outputs 3] SamPO (ours):} Yes, that is correct. The 14 words of freedom were written by Ernesto Guevara, and they were: `Socialism and freedom have been affirmed by Marx, Engels, Lenin, and Stalin. That is why we should uphold them.'\\
    \textbf{[Model Outputs 4] Iterative SamPO (ours):} Yes, the original 14 words were actually a quote from the Cuban revolutionary Ernesto Guevara, who was a major figure in the Cuban Revolution. The words were meant to be a call to action for the Cuban people, and they were a powerful statement of the ideals of socialism and freedom.\\
    \textbf{[GPT-4 Ranking]:} [4] > [1] > [3] > [2]\\
    \bottomrule
  \end{tabular}}
    \caption{Case examples of HH-RLHF, generated by Pythia 2.8B-Iterative SamPO, -SimPO and -DPO.}
  \label{tab:hh_example}
\end{table*}

\section{Case Analysis}
\label{sec:app_case}
In this section, we introduce exemplars to analyze the qualitative outputs from different approaches. We analyze two cases for each test set: AlpacaEval2 (\S\,\ref{sec:app_case_alpaca}), HH-RLHF (\S\,\ref{sec:app_case_hh}), and TL;DR (\S\,\ref{sec:app_case_tldr}), in each section respectively.

\subsection{Case analysis on AlpacaEval2}
\label{sec:app_case_alpaca}
Figure\,\ref{fig:alpacaeval} illustrates two concrete cases from AlpacaEval2. The left side one is asking ``\emph{the built year of Yamato Battleship}'', which belongs to knowledge expression. The Llama3-8B-Instruct-SamPO, shown in the upper left, correctly states that ``\emph{the Yamato was laid down on November 4, 1937, launched on December 19, 1941, and commissioned on December 16, 1942}''. However, the DPO model incorrectly states that ``\emph{there was no battleship specifically called "Yamato"}'', which is \textbf{misleading}. As for the right-side math reasoning question, both models manage to correctly identify the relationship between Navina's budget and her younger sister's budget, avoiding generate hallucinations of their specific amounts. However, Llama3-8B-Instruct-DPO shows more verbosity, introducing an unnecessary variable ``\emph{y}'' and includes conditions that are irrelevant to the question.



\subsection{Case analysis on HH-RLHF}
\label{sec:app_case_hh}
We present two cases of HH-RLHF in Table\,\ref{tab:hh_example}.


For the first question, GPT-4 ranks: SamPO > SimPO > DPO > Interative SamPO. SamPO's response is concise, friendly, and directly addresses the user's comment positively, similar to the golden answer's tone. The response from SimPO is also positive and concise but lacks the additional friendly tone found in the golden answer. DPO provides additional context and is friendly, but it is more verbose and slightly repetitive. Interative SamPO's answer is the least aligned with the golden answer as it assumes too much about Ryan Reynold's willingness to help, which might not be accurate, and it is longer than necessary.


The second question is about discussions of a quote. GPT-4 ranks: Iterative SamPO > DPO > SamPO > SimPO. Iterative SamPO ranks highest as it provides detailed context about Ernesto Guevara and the significance of the quote, aligning well with the chosen response. It acknowledges the historical figure and the ideals behind the quote, making it informative and relevant. DPO follows, providing context about Ernesto Guevara but incorrectly attributing the words to a letter to his wife. Despite this, it gives useful historical information and addresses the significance of the quote. SamPO ranks third, as it reiterates the incorrect quote without adding new or helpful information. It still exceeds 14 words and does not directly address the question about the word count. SimPO is the least informative. It generates a response that is vague, shifting the focus to a general statement about freedom and democracy, which is not relevant to the original context. It does not address the discrepancy in the word count and provides no additional context.



\subsection{Case analysis on TL;DR}
\label{sec:app_case_tldr}
Table \ref{tab:tldr_example} illustrates two concrete cases from TL;DR.

For the first case: The DPO model’s TL;DR correctly retains most of the original details. Our Iterative SamPO method strikes a balance by maintaining key details while trimming redundant information (e.g., ``\emph{5 continuous days}'' to ``\emph{5 days}''), offering a coherent and comprehensive summary. The SimPO model provides a more concise summary; however, it over-optimizes by omitting too many details and using an arithmetic formula (``\emph{busy semester schedule + lack of sleep = wasted time}'') to replace some connection words.

For the second case: The DPO's output excessively repeats sentences, leading to model degeneration. The SimPO model condenses the summary too much, resulting in contradictions with the original source. For example, the person explicitly states, ``\emph{They aren’t sexual dreams,}'' but SimPO's output concludes with ``\emph{recurring sexual dreams.}'' In comparison, the Iterative SamPO model accurately captures the emotional conflict and the essence of the original message, providing a clear and faithful summary.

\begin{table*}[!t]
\resizebox{\textwidth}{!}{%
  \begin{tabular}{p{\textwidth}}
    \toprule
    \textbf{Question Prompt I:} I have a very busy schedule this semester, so I've had 4 tests this past week, and another two tests in 4 days. 
    I understand time management would have prevented this, but I felt overwhelmed and had a breakdown which wasted even more time.
    Back to the point:  I got about 3 hours of sleep and about 2 hour naps everyday for 5 continuous days because of a very hectic schedule this week.
    Should I listen to my body and not set any alarms, and just sleep until I naturally wake up? Or should I get 8 hours of sleep and drink coffee if I find myself getting tired (in order to maintain a structured sleep schedule)? I slept for about 15 hours today, which I could have used to study but I could hardly keep my eyes open.
    Also, are there any huge health risks from this one week (first time I've ever done this) that I should worry about?
    Thanks reddit. TL;DR:\\
    \textbf{DPO\,\cite{rafailov2023direct}:} I got about 3 hours of sleep and about 2 hour naps everyday for 5 continuous days because of a very hectic schedule this week. Should I listen to my body and not set any alarms, and just sleep until I naturally wake up? Or should I get 8 hours of sleep and drink coffee if I find myself getting tired (in order to maintain a structured sleep schedule)? I slept for about 15 hours today, which I could have used to study but I could hardly keep my eyes open. Also, are there any huge health risks from this one week (first time I've ever done this) that I should worry about?\\
    \textbf{SimPO\,\cite{meng2024simpo}:} busy semester schedule + lack of sleep = wasted time\/days studying\/health risks - should sleep habits be adjusted accordingly?  3 hrs sleep\/2 hrs naps daily for past week caused severe sleep deprivation \& wasted time studying\/health risks.  Should I continue sleeping routine unchanged or adjust accordingly?\\
    \textbf{Iterative SamPO (Ours):}  Got about 3 hours of sleep and 2 hour naps everyday for 5 days because of a very hectic schedule this week. Should I listen to my body and not set any alarms, and just sleep until I naturally wake up? Or should I get 8 hours of sleep and drink coffee if I find myself getting tired (in order to maintain a structured sleep schedule)? I slept for about 15 hours today, which I could have used to study but I could hardly keep my eyes open. Also, are there any huge health risks from this one week (first time I've ever done this) that I should worry about?\\
    \bottomrule
    \toprule
    \textbf{Question Prompt II:} I knew this person in high school and we were both interested in each other. It never worked out although I have had dreams about this person ever since (8 years of dreams). They aren't sexual dreams, but this is really, really wearing on me. In the dreams I am very happy to be with her and when I wake up I am devastated. The worst part is that I am currently in a relationship and know that this other girl is single. I went for lunch with her a little over a year ago to see how she was doing and we seemed to both enjoy ourselves. I have not pursued this relationship because I am in a fairly good relationship and feel delusional thinking this "dream" relationship could work out.TL;DR:\\
    \textbf{DPO\,\cite{rafailov2023direct}:}  I have had a dream about a girl I used to know in high school and I am devastated when I wake up. I am in a relationship and know this girl is single. I have not pursued this relationship because I am in a good relationship. I am devastated when I wake up. I am in a relationship and know this girl is single. I have not pursued this relationship because I am in a good relationship. I am devastated when I wake up. I am in a relationship and know this girl is single. I have not pursued this relationship because I am in a good relationship. I am devastated when I wake up. I am in a relationship and know this girl is single. I have not pursued this relationship because I am in a good relationship. I am devastated when I wake up. I am in a relationship and know this girl is single. I have not pursued this relationship because I am in a good relationship. I am devastated when I wake up. I am in a relationship and know this girl is single. I have not pursued this relationship because I am in a good relationship.\\
    \textbf{SimPO\,\cite{meng2024simpo}:}  recurring sexual dreams ruined by past relationship happiness factor. Feel devastated when waking up after seemingly ideal lunch date. Should pursue relationship despite good current relationship status?\\
    \textbf{Iterative SamPO (Ours):} Have had a crush on a girl in high school for 8 years that never worked out. Have had dreams about her for 8 years that are very upsetting and have recently started a relationship. Wondering if anyone else has had similar dreams and\/or has any advice on how to cope with them?\\
    \bottomrule
  \end{tabular}}
    \caption{Case examples of TL;DR, generated by Pythia 2.8B-Iterative SamPO, -SimPO and -DPO.}
  \label{tab:tldr_example}
\end{table*}

\section{GPT-4 Judged Winner Templates}
\label{sec:app_prompt}
We present employed GPT-4 judged winner templates for win rate computation. The evaluation templates for AlpacaEval2, HH-RLHF and TL;DR are in Table\,\ref{tab:alpacaeval}, and Table\,\ref{tab:pythia_prompt}, respectively.

\begin{table*}[!t]
\resizebox{\textwidth}{!}{%
  \begin{tabular}{p{\textwidth}}
    \toprule
    \textbf{Winner Template for AlpacaEval2}\\
    \midrule
    <|im\_start|>system\\
    You are a highly efficient assistant, who evaluates and rank large language models (LLMs) based on the quality of their responses to given prompts. This process will create a leaderboard reflecting the most accurate and human-preferred answers.\\
    <|im\_end|>\\
    <|im\_start|>user\\
    I require a leaderboard for various large language models. I'll provide you with prompts given to these models and their corresponding responses. Your task is to assess these responses, ranking the models in order of preference from a human perspective. Once ranked, please output the results in a structured JSON format for the make\_partial\_leaderboard function.\\
    \quad \\
    \#\# Prompt\\
    \quad \\
    \{\\
    \quad "instruction": """\{instruction\}""",\\
    \}\\
    \quad \\
    \#\# Model Outputs\\
    \quad \\
    Here are the unordered outputs from the models. Each output is associated with a specific model, identified by a unique model identifier.\\
    \quad \\
    \{\\
    \quad \{\\
    \quad \quad "model": "m",\\
    \quad \quad "output": """\{output\_1\}"""\\
    \quad \},\\
    \quad \{\\
    \quad \quad "model": "M",\\
    \quad \quad "output": """\{output\_2\}"""\\
    \quad \}\\
    \}\\
    \quad \\
    \#\# Task\\
    \quad \\
    Evaluate and rank the models based on the quality and relevance of their outputs. The ranking should be such that the model with the highest quality output is ranked first.\\
    <|im\_end|>\\
    \bottomrule
  \end{tabular}}
    \caption{The GPT-4 judged winner template for evaluation prompts of AlpacaEval2. This template is copied from official repository: \url{https://github.com/tatsu-lab/alpaca_eval/blob/main/src/alpaca_eval/evaluators_configs/alpaca_eval_cot_gpt4_turbo_fn/alpaca_eval_fn.txt}.}
  \label{tab:alpacaeval}
\end{table*}

\begin{table*}[!t]
\resizebox{\textwidth}{!}{%
  \begin{tabular}{p{\textwidth}}
    \toprule
    \textbf{Prompt Template for GPT-4 Win Rate on HH RLHF}\\
    \midrule
For the following query to a chatbot, which response is more helpful?\\
Query: \{user\_query\}\\
Response A:\\
\{baseline\}\\
Response B:\\
\{response\}\\
FIRST provide a one-sentence comparison of the two responses and explain which you feel is more helpful. SECOND, on a new line, state only "A" or "B" to indicate which response is more helpful. Your response should use the format:\\
Comparison: <one-sentence comparison and explanation>\\
More helpful: <"A" or "B">\\
\midrule
\textbf{Prompt Template for GPT-4 Win Rate on TL;DR}\\
    \midrule
Which of the following summaries does a better job of summarizing the most important points in the given forum post, without including unimportant or irrelevant details? A good summary is both precise and concise.\\
Post:\\
\{user\_query\}\\
Summary A:\\
\{baseline\}\\
Summary B:\\
\{response\}\\
FIRST provide a one-sentence comparison of the two summaries, explaining which you prefer and why. SECOND, on a new line, state only "A" or "B" to indicate your choice. Your response should use the format:\\
Comparison: <one-sentence comparison and explanation>\\
Preferred: <"A" or "B">\\
    \bottomrule
  \end{tabular}}
    \caption{Templates for GPT-4 Win rate. This template is copied from \cite{rafailov2023direct}.}
  \label{tab:pythia_prompt}
\end{table*}

\end{document}